\documentclass[10pt,twocolumn,letterpaper]{article}

\usepackage[pagenumbers]{cvpr} % To force page numbers, e.g. for an arXiv version

\renewcommand{\paragraph}[1]{\vspace{.5em}\noindent\textbf{#1.}}

\usepackage{xspace}
\newcommand{\method}{MotionCrafter\xspace}

\usepackage{times}
\usepackage{epsfig}
\usepackage{graphicx}
\usepackage{amsmath}
\usepackage{amssymb}
\usepackage{multirow}
\usepackage{adjustbox}
\usepackage{fontawesome}
\usepackage{xcolor,colortbl}
\usepackage{bm}
\usepackage{nicefrac}
\usepackage{cuted}
\usepackage{smartref}
\definecolor{light}{gray}{0.95}

\def\showcomments{}    % ← 注释掉这一行即可关闭所有注释

\ifdefined\showcomments
    
    \newcommand{\chuanxia}[1]{\textcolor{teal}{[chuanxia\@: #1]}}
    \newcommand{\ruijie}[1]{{\color{blue}ruijie: #1}}
    \newcommand{\red}[1]{{\color{red}#1}}
    \newcommand{\todo}[1]{{\color{red}#1}}
    \newcommand{\TODO}[1]{\textbf{\color{red}[TODO: #1]}}
    \newcommand{\wbhu}[1]{{\color{orange}#1}}
    \newcommand{\jf}[1]{{\color{red}{[JF: #1]}}}
\else
    \newcommand{\chuanxia}[1]{}
    \newcommand{\ruijie}[1]{}
    \newcommand{\red}[1]{}
    \newcommand{\todo}[1]{}
    \newcommand{\TODO}[1]{}
    \newcommand{\wbhu}[1]{}
    \newcommand{\jf}[1]{}
\fi

\definecolor{cvprblue}{rgb}{0.21,0.49,0.74}
\usepackage[pagebackref,breaklinks,colorlinks,allcolors=cvprblue]{hyperref}

\def\paperID{***} % *** Enter the Paper ID here
\def\confName{CVPR}
\def\confYear{2026}

\title{
\method: 
Dense Geometry and Motion Reconstruction with a 4D VAE
}

\author{Ruijie Zhu$^{1,2}$ \quad Jiahao Lu$^{3}$ \quad Wenbo Hu$^{2\dagger}$\quad Xiaoguang Han$^{4}$\\
Jianfei Cai$^{5}$ \quad Ying Shan$^{2}$ \quad Chuanxia Zheng$^{1}$ \\\\
\vspace{-1.5em}
$^{1}$NTU  \quad $^{2}$ARC Lab, Tencent PCG \quad $^{3}$HKUST \quad $^{4}$CUHK(SZ) \quad $^{5}$Monash University
}

\begin{document}

\twocolumn[{
\maketitle

\begin{center}
\includegraphics[width=\linewidth]{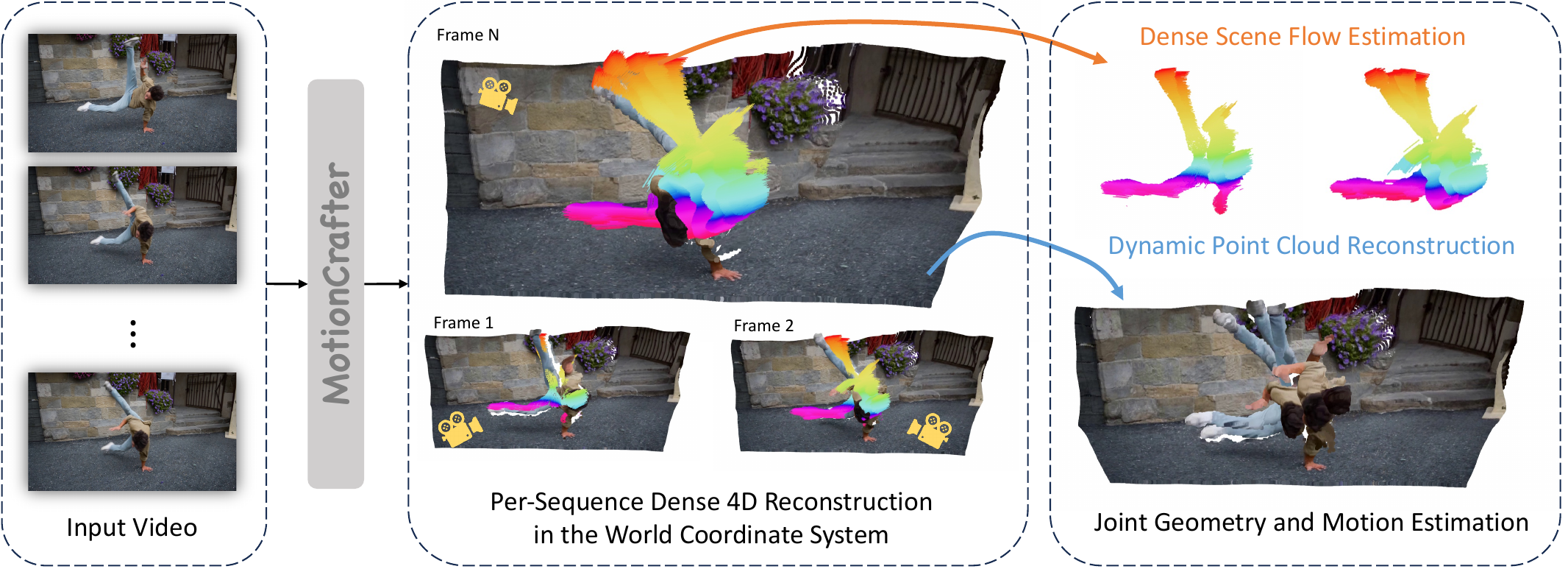}
\end{center}
\vspace{-1.0em}
\captionof{figure}{
\textbf{\method} is a video diffusion-based framework for jointly dense geometry and motion reconstruction.
Given a monocular video as input, \method simultaneously predicts dense point map and scene flow for each frame within a shared world coordinate system,
which outperforms optimization-based alternatives, yet without requiring any post-optimization.
}%
\label{fig:teaser}
\vspace{16pt}
}]
\begingroup
\renewcommand{\thefootnote}{}
% \footnotetext[0]{$^\dagger$Corresponding author.}
\endgroup

\begin{abstract}
We present \method,
a framework that leverages video generators to jointly reconstruct 4D geometry and estimate dense motion from a monocular video.
The key idea is a joint representation of dense 3D point maps and 3D scene flows in a shared coordinate system,
together with a 4D VAE tailored to learn this representation effectively.
Unlike prior work that strictly aligns 3D values and latents with RGB VAE latents—despite their fundamentally different distributions—we show that such alignment is unnecessary and can hurt performance.
Instead, we propose a new data normalization and VAE training strategy that better transfers diffusion priors and greatly improves reconstruction quality.
Extensive experiments on multiple datasets show that \method achieves state-of-the-art performance in both geometry reconstruction and dense scene flow estimation,
delivering 38.64\% and 25.0\% improvements in geometry and motion reconstruction, respectively, all without any post-optimization. 
{Project page: \url{https://ruijiezhu94.github.io/MotionCrafter_Page/}.}
\end{abstract}
\section{Introduction}
\label{sec:intro}

We consider the problem of simultaneously reconstructing \emph{4D scene geometry}
and estimating \emph{dense point motion} from a monocular RGB video of a dynamic scene 
in a feed-forward manner.
This formulation mirrors how the physical world operates: an object is structured by its geometry in 3D space,
as well as its motion across time.
Achieving this goal is highly challenging, as monocular 4D reconstruction is inherently ill-posed and dense temporal correspondences remain difficult,
especially under occlusions and significant motion.
However, a successful solution would have a wide spectrum of applications,
from video understanding to robotics~\cite{wang2017deepvo,qin2018vins} and world models~\cite{ha2018world, hafner2025mastering}.

Traditional methods tackle this problem by finding the pixel correspondences over time,
and then iteratively optimizing a 3D mesh to fit the RGB(D) observations~\cite{zollhofer2014real,newcombe2015dynamicfusion,innmann2016volumedeform}.
However, they often produce noisy results limited by the sensor,
and need per-scene optimization, which is less generalizable.
In the deep learning era, this problem is typically divided into two sub-tasks:
dynamic geometry reconstruction~\cite{zhang24monst3r,wang2025continuous,jiang2025geo4d}
and correspondence estimation~\cite{teed2020raft,karaev2024cotracker}, although they are inherently related, both relying on pixel correspondence in multi-view geometry~\cite{hartley2003multiple}.

Recent feed-forward methods such as St4RTrack~\cite{st4rtrack2025},
Dynamic Point Maps~\cite{sucar2025dynamic},
and Stereo4D~\cite{jin2024stereo4d},
have emerged as promising alternatives to address this problem,
by extending the \emph{static} 3D reconstruction networks,
like DUSt3R~\cite{wang2024dust3r} and MASt3R~\cite{leroy2024grounding},
to dynamic scenes via target-timepoint map prediction.
Even so, these methods process only \emph{pairwise} frames at once and rely on post-optimization to align the results,
reducing their ability to capture long-range motion coherence.

In this paper,
we present \method,
a framework that leverages video generators to jointly reconstruct 4D geometry and estimate dense motion
for a long monocular video sequence,
in a feed-forward manner,
\emph{without any post-optimization}.
We achieve this by proposing a \emph{world-centric} 4D representation
that denotes the dynamic scene using a sequence of point maps~\cite{wang2024dust3r,wang2025vggt,jiang2025geo4d}
and the corresponding scene flow~\cite{horn1981determining}, both defined in the world coordinate system.
This representation is intuitive and effective:
by eliminating the camera-induced motion components,
static background points ideally exhibit zero flow in the system,
making it easier to learn the motion patterns of dynamic objects.
By comparison,
prior works~~\cite{st4rtrack2025,sucar2025dynamic,jin2024stereo4d} only predict the target time point maps,
paired with the reference frame,
and do not explicitly model dense motion throughout the whole video.
We therefore argue that, to fully understand a dynamic 3D scene, it is crucial to jointly model both dense geometry and motion in a shared coordinate system throughout the \emph{entire} video sequence.

Another challenge of this task is the lack of large-scale in-the-wild datasets with dense geometry and motion.
Following recent trends in leveraging pre-trained generative models for 3D~\cite{ke2024repurposing,lu2025matrix3d,jiang2025geo4d}, we do not train our model from scratch but start from a pre-trained video generator~\cite{blattmann2023stable}.
This strategy significantly alleviates the data scarcity issue,
as the generator is trained on billions of visual data.
Moreover, the video generator inherently models spatiotemporal consistency across multiple frames,
making it well-suited for capturing long-term motion correspondence.
While Geo4D~\cite{jiang2025geo4d} has explored leveraging video generators~\cite{blattmann2023stable,xu2025geometrycrafter} for 4D reconstruction,
it only outputs \emph{independent} point maps for each frame,
without modeling dense motion across these points.

In this work, we take a further step towards jointly modeling dense geometry and motion.
We do so by encoding a unified 4D representation, combining point maps and scene flows, into a compact latent space.
Without the need to build cost volumes~\cite{teed2021raft} or establish dense correspondence~\cite{sucar2025dynamic} in pixel space, this integrated representation efficiently transfers the strong priors of the video generator to dense 4D geometry and motion reconstruction.

Moreover, we show that it is not necessary to strictly align the value range of 4D data to that of the original VAE in the diffusion model.
It is commonly believed that such alignment is essential for effectively leveraging pre-trained priors,
even for 3D geometry~\cite{ke2024repurposing,zhang2024world,lu2025matrix3d,jiang2025geo4d},
whose distribution differs substantially from that of natural images.
Our results, however, suggest otherwise.
Specifically, we adopt the canonical normalization for point maps,
\ie, centering the 3D coordinates and scaling them according to the scene's mean scale.
Despite this misalignment with the original RGB distribution in the VAE, \method{} still achieves strong generalization and accurate 4D reconstruction and motion estimation.
This finding challenges conventional beliefs and opens up new possibilities for leveraging diffusion models for geometric tasks.

To summarize, our key contributions are:
(1) We present \method,
a framework that leverages video generators to jointly reconstruct 4D scene geometry and estimate dense motion from monocular videos.
(2) We propose a novel 4D latent representation that unifies the modeling of geometry and motion, making our model simple but effective and easy to extend.
(3) We also show that strong generalization can be achieved without strictly aligning our 4D representation to the latent space of video diffusion,
challenging the conventional wisdom in diffusion-based 3D learning.

\begin{figure*}
    \centering
    \includegraphics[width=\linewidth]{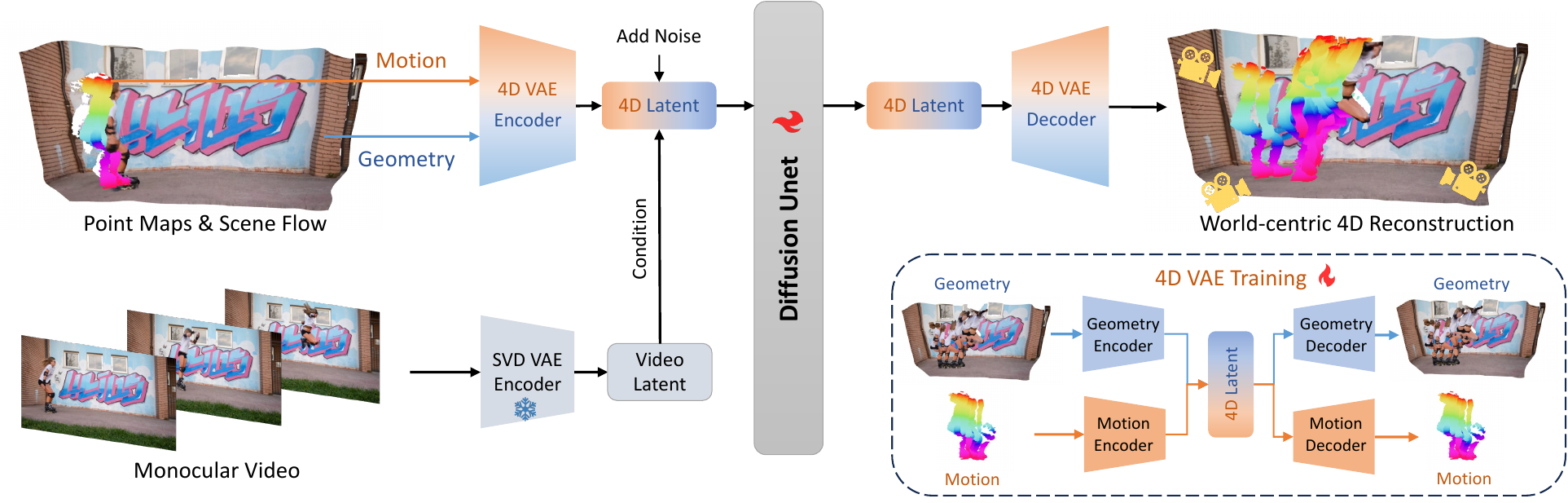}
    \caption{\textbf{Overview of \method}.
    % We define both scene geometry and motion within a shared world coordinate system. 
    We first train a novel \emph{4D VAE} (bottom-right), consisting of a \emph{Geometry VAE} and a \emph{Motion VAE}. 
    These two components jointly encode the point map and scene flow into a unified 4D latent representation. 
    Within the Diffusion Unet, we leverage the pretrained VAE from SVD (Stable Video Diffusion) to encode video latents as conditional inputs, 
    which are then channel-wise concatenated with our 4D latent to guide the denoising process. We only add noise to the 4D latents during model training for the Diffusion version.
    Note that we do not enforce the 4D latent distribution to strictly align with the original SVD VAE latent distribution.  
    And we find that this relaxed training strategy consistently improves the generalization performance of both the VAE and the Diffusion Unet.
    }
    \label{fig:overview}
    \vspace{-1em}
\end{figure*}
\section{Related Work}
\label{sec:related}

\paragraph{4D Scene Reconstruction}
Early 4D reconstruction works mainly focus on optimization-based approaches~\cite{newcombe2015dynamicfusion,pumarola2021d,li2021neural,du2021neural,tretschk2021non,park2021nerfies,fridovich2023k,cao2023hexplane,li2023dynibar,Wu_2024_CVPR,yang2024deformable,yang2024real,som2024,wang2025freetimegs},
which iteratively fits a 4D representation to monocular or multi-view videos.
With the development of neural radiance fields (NeRFs)~\cite{mildenhall2020nerf},
many time-dependent NeRFs~\cite{park2021nerfies,du2021neural,li2021neural,pumarola2021d,fridovich2023k,cao2023hexplane,li2023dynibar}
fit deformable 3D representations to dynamic scenes.
However, these approaches suffer from expensive volumetric rendering,
making them less practical for real-world applications.
3D Gaussian Splatting (3D-GS)~\cite{kerbl20233d} avoids expensive sampling using a rasterization-based rendering pipeline.
Several works~\cite{Wu_2024_CVPR,yang2024deformable,yang2024real,som2024,zhu2024motiongs,lu2024dn,wang2025freetimegs}
extend it to dynamic scene reconstruction,
which significantly reduces the rendering time.
Besides,
some methods~\cite{Zhang2022StructureAM,li2025_megasam,lu2025trackingworld} achieve accurate and robust 4D reconstruction by leveraging depth priors~\cite{hu2024-DepthCrafter,ke2024repurposing,piccinelli2024unidepth,yang2024depth,lihe24DAV2}.
However, they still require per-scene optimization.

Some recent works~\cite{zhang24monst3r,jin2024stereo4d,wang2025continuous,sucar2025dynamic,st4rtrack2025,jiang2025geo4d,xu20254dgt,AETHER,chen2025back,liang2024feed} have explored feed-forward 4D scene reconstruction from monocular videos.
Among them,
MonST3R~\cite{zhang24monst3r} adapts the notable \emph{static} 3D reconstructor DUSt3R~\cite{wang2024dust3r} to dynamic scenes.
The follow-ups~\cite{jin2024stereo4d,wang2025continuous,sucar2025dynamic,st4rtrack2025,AETHER,lu2025align3r}
took a similar path, explicitly predicting the point correspondences.
However, due to DUSt3R's limitations, they process pairs of frames at a time.
To handle long monocular videos,
$\pi^3$~\cite{wang2025pi} builds a permutation-equivariant architecture on top of VGGT~\cite{wang2025vggt} for static and dynamic 3D reconstruction.
Geo4D~\cite{jiang2025geo4d} leverages video generators~\cite{xing2024dynamicrafter} to directly infer 4D point maps from monocular videos.
4DGT~\cite{xu20254dgt} and BTimer~\cite{liang2024feed} utilize transformer-based architectures to predict dynamic 3D Gaussian representations~\cite{kerbl20233d}.
However, they do not explicitly model dense point correspondences over time.

\paragraph{Scene Flow Estimation}
Early works define point correspondences as optical flow estimation in pixel space~\cite{horn1981determining,brox2004high,bruhn2005lucas,ilg2017flownet,teed2020raft}.
One popular pipeline is to directly estimate dense pixel-wise correspondences in a coarse-to-fine manner~\cite{dosovitskiy2015flownet,ilg2017flownet,sun2018pwc,teed2020raft}.
However, such a coarse-to-fine strategy may fail in the presence of large motions or occlusions~\cite{revaud2015epicflow}.
More recently,
GMFlow~\cite{Xu_2022_CVPR,xu2023unifying} proposes to reformulate optical flow estimation as a global matching problem rather than local regression.
The follow-ups~\cite{huang2022flowformer,shi2023flowformer++} also use transformer-based neural networks to model the global correlations.
% As for the long-range point tracking~\cite{rubinstein2012towards,sand2008particle},
% recent works~\cite{doersch2022tap,karaev2024cotracker,karaev2025cotracker3} explore to learn robust point trackers from large-scale video datasets.
However, these methods still deal with 2D point correspondences in image space.
In contrast,
we address 3D scene flow estimation in world space.
Some researchers have also explored estimating 3D \emph{scene flow} directly from image pairs,
including RAFT-3D~\cite{teed2021raft}, SpatialTracker~\cite{SpatialTracker}, SceneTracker~\cite{wang2025scenetracker}, and TAPVid-3D~\cite{koppula2024tapvid}.
More closely related to our work,
several recent works~\cite{sucar2025dynamic,st4rtrack2025,jin2024stereo4d} explore reconstructing dynamic 3D geometry,
along with 3D scene flow estimation in world space.
However, they process only two images at a time and require post-processing to refine the results.

\paragraph{Geometric Diffusion Model}
Like our approach,
many recent works have leveraged pre-trained off-the-shelf diffusion models~\cite{rombach2022high,ho2022video,wang2023modelscope,singer2023makeavideo,blattmann2023align,ge2023preserve,wang2023videocomposer,guo2024animatediff,blattmann2023stable,zhang2024show,huang2024blue,xing2024dynamicrafter,kong2024hunyuanvideo,liang2025diffusion,garcia2025fine}
to tackle 3D tasks~\cite{gao2024cat3d,sargent2024zeronvs,chen2024mvsplat360,yu2024viewcrafter,szymanowicz2025bolt3d,li2025dso,wu2025amodal3r,ke2024repurposing,ye2024stablenormal,xu2024matters,song2025depthmaster,liu2023zero,shi2024mvdream,zheng2024free3d,voleti2025sv3d}, thanks to rich priors learned from large-scale image or video datasets.
When it comes to 4D reconstruction, a straightforward solution is to generate multi-view videos,
and then fit a 4D representation via per-scene optimization~\cite{xie2024sv4d,zeng2024stag4d,wu2025cat4d}.
Inspired by score distillation sampling (SDS)~\cite{pooledreamfusion},
another line of works~\cite{jiang2024consistentd,zhang20244diffusion,dreamscene4d,ren2023dreamgaussian4d,li2024dreammesh4d}
directly distill 4D priors from pre-trained video generators.
However, these approaches still rely on iterative per-scene optimization,
which is expensive when dealing with in-the-wild videos.
The most related work is Geo4D~\cite{jiang2025geo4d} that fine-tunes a pre-trained video diffusion model to directly infer dynamic 3D point maps, depths, and camera poses from monocular videos.
Our method differs from Geo4D in two aspects:
(1) we \emph{simultaneously} reconstruct \emph{dynamic 3D geometry} and estimate \emph{dense point correspondences} in a unified 4D VAE framework; and
(2) we show that it is \emph{not} necessary to align the data and latent spaces when fine-tuning a diffusion model.
\section{Method}
\label{sec:method}

Given as input a monocular video sequence with dynamic objects,
our goal is to learn a neural network $f_\theta$ that can output a 4D representation of its geometry along with dense point-wise correspondences, \emph{simultaneously}:
\begin{equation}
    \label{eq:goal}
    f_\theta: \{\bm{I}_i\}_{i=1}^N \rightarrow
    \{\bm{X}_i, \bm{V}_{i\rightarrow i+1}\}_{i=1}^N.
\end{equation}
$\mathcal{I}=\{\bm{I}_i\}_{i=1}^N$ is the input monocular video sequence with $N$ frames,
where each frame
$\bm{I}_i\in\mathbb{R}^{H\times W\times 3}$
is an RGB image.
The network
$f_\theta$
predicts a viewpoint-invariant point map
$\bm{X}_i\in\mathbb{R}^{H\times W\times 3}$ for each frame $i$,
and the 3D scene flow\footnote{Unless otherwise noted,
we simplify $\bm{V}_{i\rightarrow i+1}$ as $\bm{V}_{i}$ for clarity.}
$\bm{V}_{i\rightarrow i+1}\in\mathbb{R}^{H\times W\times 3}$
between each pair of consecutive frames $i$ and $i+1$.
Both the point map and scene flow are represented in a shared \emph{world coordinate system}.
Note that,
since we only predict forward scene flow,
the last frame
$N$
does not have a corresponding flow prediction,
i.e., we do not supervise $\bm{V}_{N\rightarrow N+1}$.

To smoothly model the long-term motion and enable generalization to diverse scenes,
we build $f_\theta$ upon a pretrained video generator,
where $\boldsymbol{\theta}$ denotes the learnable parameters.
Our framework is illustrated in~\cref{fig:overview}.
We first introduce our unified 4D representation in~\Cref{sec:method_pre}.
Then,
in~\Cref{sec:method_vae},
we present a dedicated 4D\footnote{Here we use 4D VAE to refer to the fused geometry and motion VAEs.} VAE architecture that jointly encodes geometry and motion into a unified latent space.
Finally,
in~\Cref{sec:method_training},
we describe the overall training and inference strategy for our model.

\subsection{Unified Geometry \& Motion Representation}
\label{sec:method_pre}

\begin{figure}
    \centering
    \includegraphics[width=\linewidth]{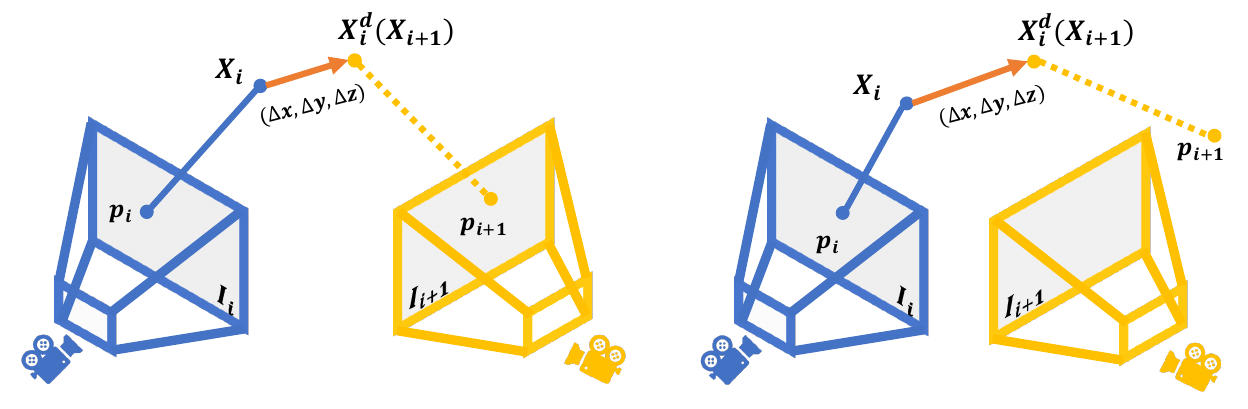}
    \captionof{figure}{\textbf{Geometry and Motion representation.}
    % We define both point maps and scene flows in the world coordinate system.
    For a pixel $p_i$ in frame $\bm{I}_i$,
    $\bm{X}_i$ is its corresponding 3D point.
    As this 3D point moves,
    we use $\bm{X}_i^d$ to represent the moved point and $\bm{V}_{i} = (\Delta x, \Delta y, \Delta z)$ to represent the motion.
    Ideally,
    $\bm{X}_i^d$ should align with a matching point $\bm{X}_{i+1}$ in next frame $\bm{I}_{i+1}$.
    However, their pixel indexes are totally different ($p_i$ vs. $p_{i+1}$) and $p_{i+1}$ might even be out of view due to camera/object motion,
    making it impossible to build one-to-one correspondence between $\bm{X}_i^d$ and $\bm{X}_{i+1}$.
    }%
    \label{fig:representation}
    \vspace{-1em}

\end{figure}

Here,
like in DUSt3R~\cite{wang2024dust3r},
we define the point maps and scene flows in the coordinate system of the first frame,
which serves as the world coordinate system.
In particular,
the point map
$\bm{X}_i \in \mathbb{R}^{H\times W\times 3}$
stores the 3D coordinates
$(x, y, z)$
of each pixel from the frame
$i$
in the world coordinate system,
while the scene flow
$\bm{V}_{i} \in \mathbb{R}^{H\times W\times 3}$
represents the 3D motion vector \((\Delta x, \Delta y, \Delta z)\)
of each pixel from the frame
$i$
to
$i+1$.
Ideally, the \emph{deformed point map}
\begin{equation}
    \bm{X}_i^{d} = \bm{X}_i + \bm{V}_i
\end{equation}
should be spatially aligned with the point map of the next frame
$\bm{X}_{i+1}$. 
However,
due to viewpoint changes,
$\bm{X}_i^{d}$
and
$\bm{X}_{i+1}$
are not in one-to-one correspondence in pixel space, 
as they represent different frame contents,
as illustrated in~\cref{fig:representation}.
Note that,
% following the very recent work~\cite{sucar2025dynamic,st4rtrack2025},
our scene flow is also defined directly in the \emph{(world) coordinate system},
meaning that each
$\bm{V}_i$
represents the motion vector \((\Delta x, \Delta y, \Delta z)\)
in the world space,
naturally eliminating camera-induced motion components.

Such a unified geometry-motion representation offers several advantages:
%  for both modeling and learning:
1) \emph{Camera-free modeling.} 
Similar to DUSt3R~\cite{wang2024dust3r},
defining the geometry and motion in a chosen world coordinate system removes the need for additional camera pose estimation.
2) \emph{Temporal consistency.} 
In a continuous video sequence, geometry and motion are temporally coherent. 
Modeling them jointly in the same coordinate system makes them easier to learn.
3) \emph{Richer motion modeling.}
Unlike existing methods~\cite{sucar2025dynamic,st4rtrack2025},
we define scene flow between every pair of consecutive frames in the video,
rather than only between the first frame and others.
Consequently, this representation is less sensitive to occlusions induced by viewpoint variations and remains capable of capturing motion information of newly emerging dynamic objects in subsequent frames.

\subsection{Unified 4D Geometry-Motion VAE}
\label{sec:method_vae}

Here, we describe how to effectively encode the above 4D representation into a latent space,
which can then be used as the target for a video generator.
Recent works on geometric diffusion models~\cite{ke2024repurposing,zhang2024world,jiang2025geo4d} encode only 3D geometric attributes, neglecting explicit motion modeling for dynamic scenes.
In contrast, we design a novel 4D VAE architecture that jointly encodes geometry and motion into a unified 4D latent,
as illustrated in~\cref{fig:overview}.

To leverage the priors of \emph{pretrained} video generators,
it is widely believed that
\emph{the input to the VAE should be strictly aligned with the original data distribution of the pre-trained diffusion model}~\cite{ke2024repurposing, zhang2024world, jiang2025geo4d}.
That is,
a naive approach is to directly rescale 3D attributes (such as disparity~\cite{ke2024repurposing} and point maps~\cite{zhang2024world}) into the range $[-1, 1]$,
by performing a max normalization,
and then encode them using the frozen VAE weights.
However,
the world-coordinate 3D attributes are typically unbounded, with coordinates spanning $(-\infty, +\infty)$,
in contrast to images with bounded pixel ranges $[0, 255]$.
Moreover, the distribution of 3D attributes is inherently distinct from that of natural RGB images.
Hence,
in this work,
we investigate a fundamental question:
\emph{Is strict alignment with the diffusion model's input space essential for finetuning diffusion models?}

\begin{figure}
    \centering
    \includegraphics[width=\linewidth]{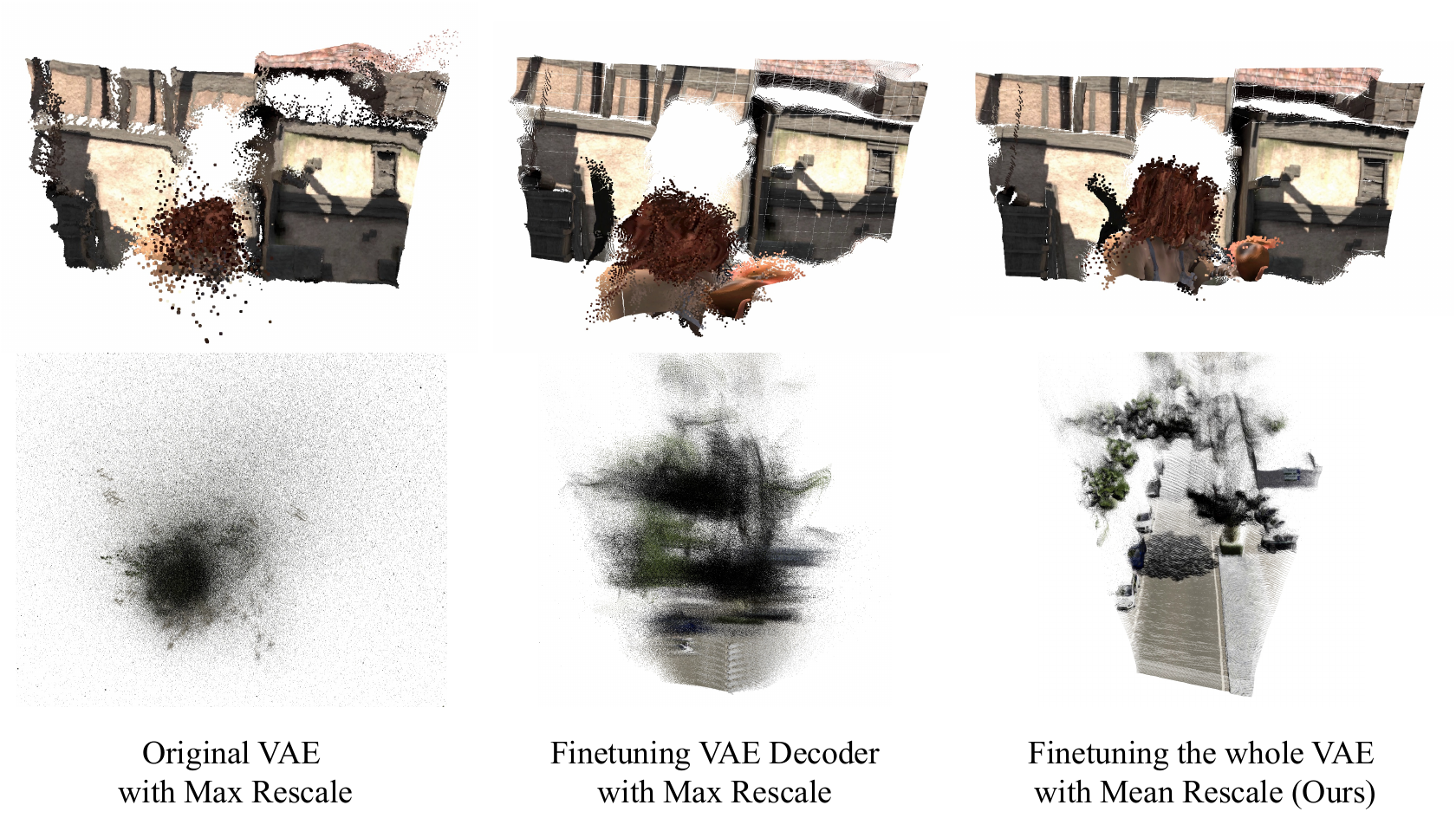}
    \caption{\textbf{Results of different normalization and VAE training strategies}.
    For outdoor scenes with significant variations in depth (the second row), 
    the \emph{original VAE} fails to recover the scene structure.
    Even with decoder fine-tuning,
    the reconstruction quality remains poor.
    Our proposed \emph{mean} normalization and VAE training strategy significantly improve reconstruction quality.
    }
    \label{fig:rescale}
    \vspace{-1em}
\end{figure}

\paragraph{Geometry VAE with Revised Normalization}
To answer the above question,
one key insight of our model is a slightly adjusted point map normalization strategy for the Geometry VAE.
Note that,
unlike the max normalization to $[-1, 1]$ commonly used in existing geometric diffusion models~\cite{ke2024repurposing,zhang2024world,jiang2025geo4d},
we instead apply \emph{canonical normalization} to each sequence of world-coordinate point maps:
\begin{equation}
    \label{eq:normalization}
    \hat{\bm{X}}_i = \frac{\bm{X}_i-\mu}{S},
\end{equation}
where
$\mu=\frac{1}{|\mathcal{D}|}\sum_{d\in\mathcal{D}} \bm{X}_d$
is the mean of all valid points,
denoted by
$\mathcal{D}$,
in the point map sequence
$\cup_{i=1}^N\{\bm{X}_i\}$
and
$S=\frac{1}{|\mathcal{D}|}\sum_{d\in\mathcal{D}} \big\| \bm{X}_d - \mu \big\|_2 + \varepsilon$ 
is the mean distance for scale normalization with a small constant
$\varepsilon$
for numerical stability.
This normalization maintains the scale invariance of point maps,
while significantly improving the reconstruction quality of the Geometry VAE and better preserving finer structural details compared to max normalization~\cite{wang2024dust3r,wang2025vggt},
especially when handling large-scale outdoor scenes, as shown in~\cref{fig:rescale}.

Here, we finetune the entire encoder-decoder using our new normalization strategy, which provides a more flexible input distribution.
We define the training objective as:
\begin{equation}
\label{eq:geometry_loss}
\mathcal{L}_G = \mathcal{L}_{\text{point}} + \lambda_{d}\mathcal{L}_{\text{depth}} + \lambda_{n}\mathcal{L}_{\text{normal}},
\end{equation}
where
$\mathcal{L}_{\text{point}}$
is the MSE loss for point map reconstruction,
$\mathcal{L}_{\text{depth}}$
is a multi-scale loss computed on the projected depth maps,
and
$\mathcal{L}_{\text{normal}}$ enforces consistency of surface normals, following ~\cite{xu2025geometrycrafter, wang2024moge}.
The difference is that we are encoding world-based point clouds.
Hence, we normalize the ground-truth camera poses together with the point clouds so that
we can use scale-aligned camera parameters to project the point clouds into depth maps.
Experiments show that this supervision, similar to the multimodal fusion in Geo4D~\cite{jiang2025geo4d}, improves the reconstruction quality of point clouds.

Here we also tried using Kullback–Leibler (KL) divergence loss~\cite{kullback1951information} to constrain the distribution of the latent to a standard Gaussian distribution,
but found that it led to a significant drop in VAE performance.

The proposed normalization and VAE training strategy consistently improve the performance of the VAE and the downstream diffusion U-Net in~\cref{tab:ablation_geometry},
suggesting that \emph{strict alignment with the diffusion model's input and latent space is not always necessary},
especially for the 3D attributes. 

\paragraph{Motion VAE}
A simple way to model scene flow is to train a separate motion VAE with the same architecture as the Geometry VAE.
However, because motion and geometry are inherently correlated,
learning motion independently may be suboptimal.
We therefore explore several fusion strategies between geometry and motion:
(1) \textit{no fusion}, where geometry and motion are encoded separately without any interaction;
(2) \textit{offset fusion}, inspired by LayerDiffuse~\cite{zhang2024layerdiffuse},
where the motion latent is added as an offset to the geometry latent; and
(3) \textit{unified fusion}, where the geometry and motion latents are concatenated into a unified 4D latent
and passed to the motion VAE decoder to reconstruct the scene flow.
As reported in~\cref{tab:ablation_motion}, although the unified concatenation strategy does not yield the best reconstruction quality at the VAE stage,
it leads to superior performance in the subsequent diffusion U-Net.
During Motion VAE training, we freeze the Geometry VAE's parameters to preserve its learned geometric priors.
The training objective is formulated as:
% \begin{equation}
%     \label{eq:motion_loss}
%     \mathcal{L}_{M} = \mathcal{L}_{\text{sceneflow}} + \lambda_{\text{reg}}\mathcal{L}_{\text{reg}},
% \end{equation}
% \begin{equation}
%     \label{eq:motionreg_loss}
%     \mathcal{L}_{\text{reg}} = ,
% \end{equation}
\begin{equation}
\label{eq:motion_loss}
\mathcal{L}_{\text{M}} = 
\underbrace{
\frac{1}{|\mathcal{D}|} \sum_{d \in \mathcal{D}} \, \| \hat{\bm{V}}_d - \bm{V}_d \|_2^2
}_{\text{Scene flow reconstruction loss}}
+ \lambda_{\text{reg}} \underbrace{
\frac{1}{|\mathcal{N}|} \sum_{n \in \mathcal{N}} \, \| \hat{\bm{V}}_n \|_2^2
}_{\text{Zero-flow regularization}},
\end{equation}
where $\hat{\bm{V}}_d$ is the predicted scene flow, ${\bm{V}}_d$ is the ground truth,
$\mathcal{D}$ denotes valid pixels,
and $\mathcal{N}$ denotes all pixels.
The first term denotes the MSE loss on the valid scene flow,
and the second term is a regularization term to encourage the scene flow to zero,
following the as-static-as-possible assumption.

By combining the Geometry VAE and Motion VAE into a unified 4D VAE, 
we successfully achieve an integrated representation of geometry and motion within a single latent space, 
enabling efficient 4D scene encoding and decoding.

\subsection{Model Training}
\label{sec:method_training}

\paragraph{Training Data}
Dynamic datasets with annotated 3D geometry and dense scene flow are difficult to collect in real-world settings.
Therefore,
we rely on synthetic datasets for training the scene flow estimation task.
In particular,
we divide our training data into two categories:
(1) \emph{Geometry Datasets:}
Dynamic Replica~\cite{karaev2023dynamicstereo},
GTA-SFM~\cite{wang2020gtasfm},
MatrixCity~\cite{li2023matrixcity},
MVS-Synth~\cite{huang2018deepmvs},
Point Odyssey~\cite{zheng2023pointodyssey},
TartanAir~\cite{wang2020tartanair},
ScanNet++~\cite{yeshwanth2023scannet++},
BlinkVision~\cite{li2024blinkvision},
OmniWorld~\cite{zhou2025omniworld}
and Synthia~\cite{ros2016synthia};
(2) \emph{Geometry-and-Motion Datasets:}
Kubric~\cite{greff2021kubric},
Spring~\cite{Mehl2023_Spring},
and Virtual KITTI 2~\cite{cabon2020virtual}.
The first category provides only geometric data,
including per-frame depth maps,
camera intrinsics and extrinsics.
Following DUSt3R~\cite{wang2024dust3r},
we express the ground-truth point clouds into a shared first frame coordinate system.
The second category additionally provides dense scene flow annotations.
During \textit{geometry reconstruction} training,
we use both dataset groups (1)+(2),
whereas during \textit{motion reconstruction} training,
only datasets in group (2) are employed.

\paragraph{Training Strategy}
We adopt a two-stage training pipeline for the VAE components.
We begin by training the Geometry VAE independently to capture scene geometry.
Next,
we train the Motion VAE while keeping the Geometry VAE frozen,
thereby preserving its learned geometric priors.
After convergence,
we combine them into a unified 4D VAE,
whose parameters remain frozen during the training of the diffusion U-Net.
For U-Net training, we combine the datasets from groups (1) and (2) to provide geometry supervision, and use only the datasets from group (2) for motion supervision.
Following prior works~\cite{hu2024-DepthCrafter,xu2025geometrycrafter} in employing EDM~\cite{karras2022elucidating} pre-conditioning,
our framework supports both the \textit{deterministic} and \textit{denoising} paradigms.
For the deterministic paradigm, the training objective is defined as:
\begin{equation}
\mathcal{L}_{\text{deterministic}} = \mathcal{L}_{\text{latent}} + \lambda_G \mathcal{L}_G + \lambda_M \mathcal{L}_{\text{M}},
\end{equation}
where $\mathcal{L}_G$ is the geometry reconstruction loss defined in~\eqref{eq:geometry_loss},
$\mathcal{L}_{M}$ is the motion reconstruction loss defined in~\eqref{eq:motion_loss}, and $\mathcal{L}_{\text{latent}}$ denotes the latent-space diffusion loss:
\begin{equation}
\mathcal{L}_{\text{latent}} =
\underbrace{
\frac{1}{N} \sum_{N} 
\| 
\hat{\mathbf{z}}^{\text{G}}_i - \mathbf{z}^{\text{G}}_i
\|_2^2
}_{\text{geometry latent supervision}}
+
\underbrace{
\frac{1}{N-1} \sum_{N-1}
\| 
\hat{\mathbf{z}}^{\text{M}}_i - \mathbf{z}^{\text{M}}_i
\|_2^2
}_{\text{motion latent supervision}},
\end{equation}
where $N$ denotes the number of frames,
and $\hat{\mathbf{z}}^{\text{G}}_i,\mathbf{z}^{\text{G}}_i,\hat{\mathbf{z}}^{\text{M}}_i, \mathbf{z}^{\text{M}}_i$ are the denoised latent and the original latent of geometry and motion, respectively.
We only perform forward scene flow estimation,
thus discarding the motion latent in the last frame.
For the denoising paradigm, the objective simplifies to latent supervision:
% \begin{equation}
$\mathcal{L}_{\text{denoise}} = \mathcal{L}_{\text{latent}}$.
% \end{equation}
We found experimentally that the deterministic paradigm generally performs better, so we use it by default. Ablation experiments can be found in the supplementary materials.

This progressive and modular training pipeline allows the model to first acquire strong geometry and motion priors before integrating temporal reasoning,
ultimately enabling robust and coherent dense 4D reconstruction.

\paragraph{Implementation Details}
To inherit the strong priors from the video generator,
both the VAE and U-Net of our \method{} are initialized with the pretrained weights of SVD~\cite{blattmann2023stable},
and trained using the AdamW optimizer~\cite{loshchilov2017decoupled} with a learning rate of $1e$-$4$. 
We first train the Geometry VAE for 40{,}000 iterations,
followed by training the Motion VAE for 20{,}000 iterations. 
Subsequently,
we merge the Geometry VAE and Motion VAE into a unified 4D VAE,
and train the U-Net for another 40{,}000 iterations with the encoded 4D latent representations.
The batch size is set to 8 for VAE training and 25 for U-Net training.
All experiments are conducted on 8 GPUs with 40 GB of memory each and take about 3 days.
More implementation details are provided in the supplementary material.

\begin{table*}[t]
\centering
\caption{
\textbf{Evaluation on joint world-centric geometry and motion reconstruction.}
All metrics are reported without percentage symbols for readability.
% Lower values of $\text{Rel}^p$ and EPE indicate better accuracy, while higher $\delta^p$ and APD denote stronger performance. 
* denotes not zero-shot scene flow evaluation.
-S and -P denote the Sequence mode and Pair mode of ST4RTrack.
Since ST4RTrack always compares with the first frame for motion,
for a fair comparison, we run it on every pair of consecutive frames and then transform the results into the world coordinate system using VGGT poses.
Plus, we add Zero-MSF + GT pose as a reference.
}

% \vspace{-0.5em}
\label{tab:joint_geo_motion}
\begin{adjustbox}{width=\textwidth}
\begin{tabular}{@{}l cc cc cc cc cc c@{}}
\toprule
% \multirow{2}{*}{\textbf{Task}} & 
\multirow{1}{*}{\textbf{Method}} &
\multicolumn{2}{c}{\textbf{Kubric}~\cite{greff2021kubric}} &
\multicolumn{2}{c}{\textbf{Spring}~\cite{Mehl2023_Spring}} &
\multicolumn{2}{c}{\textbf{VKITTI2}~\cite{cabon2020virtual}} &
\multicolumn{2}{c}{\textbf{Dynamic Replica}~\cite{karaev2023dynamicstereo}} &
\multicolumn{2}{c}{\textbf{Point Odyssey}~\cite{zheng2023pointodyssey}} &
\textbf{Avg.} \\
\cmidrule(lr){2-3} \cmidrule(lr){4-5} \cmidrule(lr){6-7} \cmidrule(lr){8-9} \cmidrule(lr){10-11}
\ \underline{\emph{Geometry}} &
$\text{Rel}^{p}\!\downarrow$ & $\delta^{p}\!\uparrow$ &
$\text{Rel}^{p}\!\downarrow$ & $\delta^{p}\!\uparrow$ &
$\text{Rel}^{p}\!\downarrow$ & $\delta^{p}\!\uparrow$ &
$\text{Rel}^{p}\!\downarrow$ & $\delta^{p}\!\uparrow$ &
$\text{Rel}^{p}\!\downarrow$ & $\delta^{p}\!\uparrow$ &
\textbf{Rank} $\!\downarrow$ \\
\cmidrule(lr){2-12}
% \multirow{7}{*}{\centering\rotatebox[origin=c]{90}{Geometry}} 
% \ \underline{\emph{Geometry}} & & & & & & & & & & & \\
POMATO~\cite{zhang2025pomato} + VGGT & 25.56 & 77.85 & 98.13 & 61.71 & 32.16 & 54.89 & 7.26 & 94.64 & 19.88 & 80.08 & 5.0 \\
ST4RTrack-S~\cite{st4rtrack2025} + VGGT & 6.61 & 95.59 & 123.27 & 42.26 & 67.68 & 21.87 & 5.65 & 96.29 & 31.00 & 68.17 & 4.0 \\
ST4RTrack-P~\cite{st4rtrack2025} + VGGT & 17.81 & 80.76 & 157.05 & 38.00 & 84.77 & 14.46 & 4.87 & 97.13 & 29.13 & 71.66 & 3.4 \\
DELTA~\cite{ngo2024delta} + VGGT & 14.09 & 85.73 & 106.88 & 50.81 & 57.03 & 42.76 & 23.21 & 74.35 & 51.47 & 50.92 & 6.0 \\
Zero-MSF~\cite{liang2025zero} + VGGT & 8.79 & 94.73 & 142.44 & 40.66 & 15.76 & 80.04 & 7.11 & 97.03 & 22.55 & 78.27 & 4.6 \\
\textcolor{gray}{Zero-MSF~\cite{liang2025zero} + GT} 
& \textcolor{gray}{8.78} & \textcolor{gray}{94.73} 
& \textcolor{gray}{142.45} & \textcolor{gray}{40.69} 
& \textcolor{gray}{11.22} & \textcolor{gray}{89.92} 
& \textcolor{gray}{7.11} & \textcolor{gray}{97.01} 
& \textcolor{gray}{22.52} & \textcolor{gray}{78.27} 
& - \\
\method (ours) & \textbf{3.40} & \textbf{98.73} & \textbf{29.20} & \textbf{77.27} & \textbf{14.60} & \textbf{84.58} & \textbf{4.04} & \textbf{99.00} & \textbf{9.94} & \textbf{94.90} & \textbf{1.0} \\
\midrule
\midrule
\ \underline{\emph{Motion}} & 
EPE$\!\downarrow$ & APD$_{0.05}\!\uparrow$ &
EPE$\!\downarrow$ & APD$_{0.1}\!\uparrow$ &
EPE$\!\downarrow$ & APD$_{0.3}\!\uparrow$ &
EPE$\!\downarrow$ & APD$_{0.05}\!\uparrow$ &
EPE$\!\downarrow$ & APD$_{0.05}\!\uparrow$ &
\\
\cmidrule(lr){2-12}
% \multirow{7}{*}{\centering\rotatebox[origin=c]{90}{Motion}}
% \underline{\emph{Motion}} & & & & & & & & & & & \\
POMATO~\cite{zhang2025pomato} + VGGT & 79.58 & 5.23 & 180.13 & 20.16 & 368.78 & 15.66 & 14.56* & 36.34* & 31.94* & 30.32* & 5.0 \\
ST4RTrack-S~\cite{st4rtrack2025} + VGGT & 59.34* & 1.68* & 105.02 & 33.47 & 156.85 & 22.11 & 1.12* & 97.13* & 10.37* & 53.39* & 4.0 \\
ST4RTrack-P~\cite{st4rtrack2025} + VGGT & 217.20* & 5.65* & 441.84 & 9.07 & 874.94 & 13.16 & 0.99* & 97.55* & 58.62* & 51.19* & 4.8 \\
DELTA~\cite{ngo2024delta} + VGGT & 8.29* & 52.95* & 8.59 & 81.55 & 156.64 & 17.7 & 0.75 & 99.57 & 6.09 & 62.11 & 3.2 \\
Zero-MSF~\cite{liang2025zero} + VGGT & 8.59* & 50.13* & 7.78* & 85.59* & 112.99* & 21.69* & 0.59* & \textbf{99.80}* & 3.58* & 80.12* & 2.4 \\
\textcolor{gray}{Zero-MSF~\cite{liang2025zero} + GT} & \textcolor{gray}{5.74*} & \textcolor{gray}{72.37*} & \textcolor{gray}{5.50*} & \textcolor{gray}{87.59*} & \textcolor{gray}{73.81*} & \textcolor{gray}{25.15*} & \textcolor{gray}{0.40*} & \textcolor{gray}{99.80*} & \textcolor{gray}{2.30*} & \textcolor{gray}{91.04*} & - \\
\method (ours) & \textbf{4.60}* & \textbf{68.01}* & \textbf{5.61}* & \textbf{90.17}* & \textbf{71.75}* & \textbf{25.90}* & \textbf{0.51} & 99.72 & \textbf{3.49} & \textbf{80.66} & \textbf{1.0} \\
\bottomrule
\end{tabular}
\end{adjustbox}
% \vspace{-1em}
\end{table*}

\begin{figure*}
    \centering
    \includegraphics[width=\linewidth]{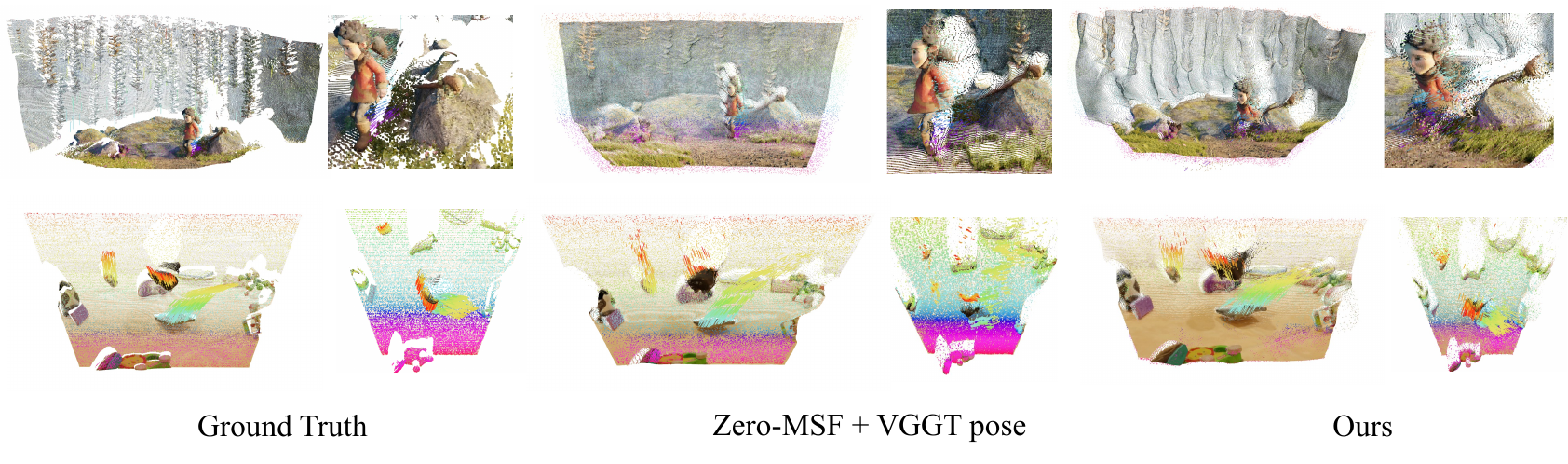}
    \caption{\textbf{Qualitative comparison with Zero-MSF~\cite{liang2025zero}.} Zoom in for the details. Compared to Zero-MSF, we have a more reasonable scene structure and better geometric details. More importantly, our predicted 3D scene flow has a more accurate direction of motion.}
    \label{fig:indomain_result}
    \vspace{-1em}
\end{figure*}

\section{Experiments}
\label{sec:experiments}
\subsection{Evaluation Setting}

\paragraph{Datasets}
For \emph{geometry} evaluation, we perform zero-shot testing on three unseen dynamic scene datasets:
DDAD~\cite{ddad2020},
Monkaa~\cite{mayer2016large},
and Sintel~\cite{sintel}. 
These datasets cover both real-world and synthetic scenes,
including indoor and outdoor environments. 
For \emph{motion} evaluation,
due to the limited availability of datasets with dense scene flow annotations,
we use a combination of three in-domain datasets
(Kubric~\cite{greff2021kubric},
Spring~\cite{Mehl2023_Spring},
and VKITTI2~\cite{cabon2020virtual})
and two out-of-domain datasets
(Dynamic Replica~\cite{karaev2023dynamicstereo}
and Point Odyssey~\cite{zheng2023pointodyssey}). 
Since Dynamic Replica and Point Odyssey only provide sparse scene flow annotations,
we compute metrics only on the annotated points.

\paragraph{Metrics}
Unlike previous methods,
we evaluate geometry and motion in the \textit{world coordinate system}.
% to directly assess the real reconstruction ability of the model. 
The predicted world-space point cloud is aligned with the ground truth by optimizing per-sequence scale and shift parameters. 
We report the \textit{relative point error} ($\text{Rel}^p$) and the \textit{percentage of inlier} ($\delta^p$, threshold 0.25) as evaluation metrics. 
The predicted scene flow is aligned according to the point map scale.
We compute the \textit{End Point Error} (EPE) and the \textit{Average Percent of Points within Delta} (APD), 
where the subscript of APD denotes the inlier threshold in the metric scale. 
We provide details in the supplementary material.

\subsection{Comparison with the State-of-the-art Methods}

\begin{figure}
    \vspace{-0.5em}
    \centering
    \includegraphics[width=\linewidth]{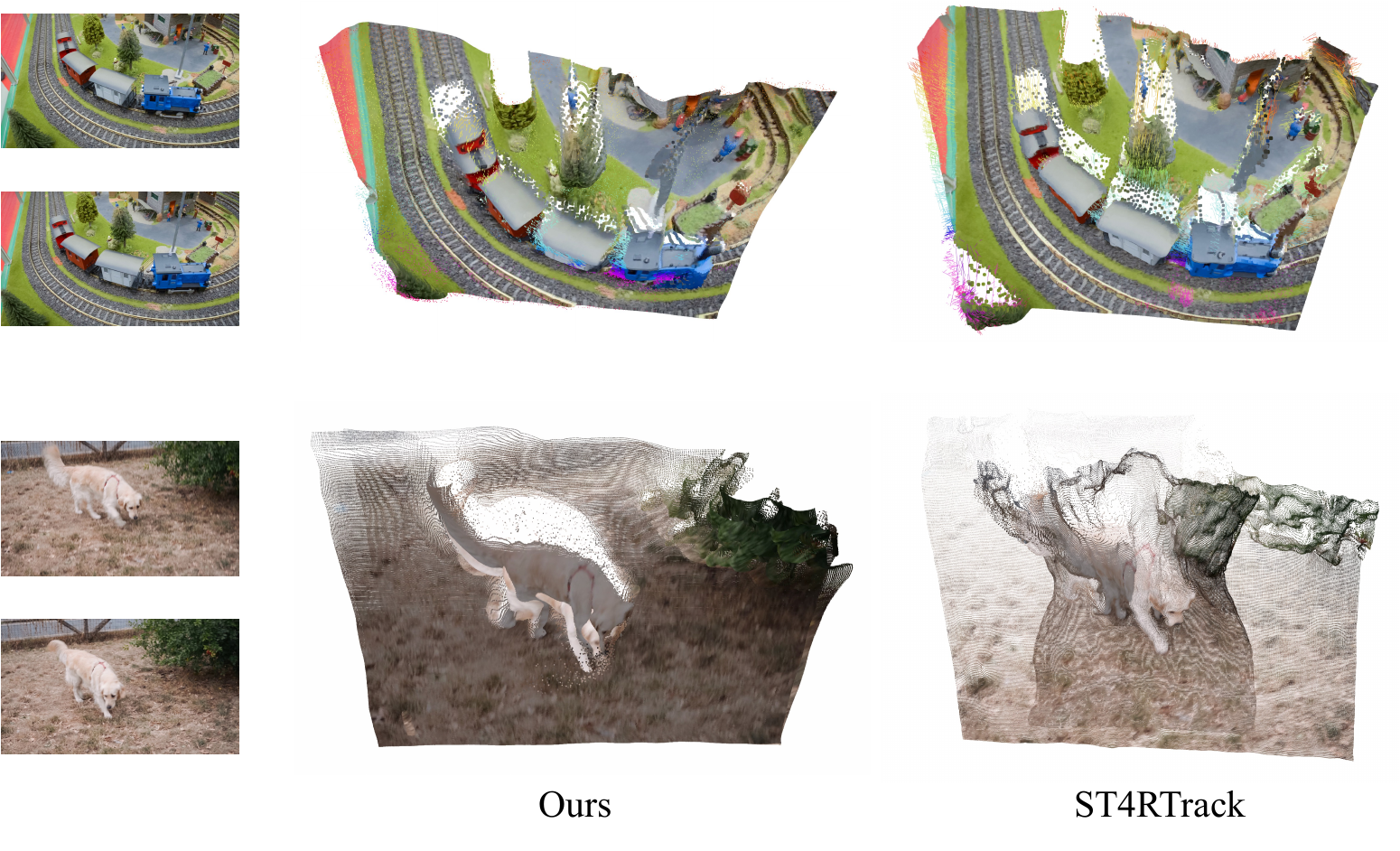}
    \caption{\textbf{Qualitative comparison with ST4RTrack~\cite{st4rtrack2025}.}
    In the first case, the pixel trajectory shows that we yield cleaner scene flow, while ST4RTrack suffers from noisy drift.
    In the second case, the deformed point map (with darker color) shows that our method predicts more temporally consistent geometry and motion.
    }
    \label{fig:effectiveness}
    \vspace{-0.5em}
\end{figure}

\paragraph{Assessing joint Geometry and Motion Reconstruction}
We compare \method with recent representative methods in joint geometry and motion estimation in~\cref{tab:joint_geo_motion}.
Here,
we assess these metrics in the world coordinate space.
Our model directly outputs a sequence of predictions in world coordinates.
By contrast, most existing methods follow a pairwise design based on DUSt3R~\cite{wang2024dust3r},
which need post-optimization or camera poses to align with ground truth.
To ensure fairness, we use the camera pose predicted by VGGT~\cite{wang2025vggt} to transform their predictions into the world coordinate system.
Through both quantitative and qualitative comparison in~\cref{fig:indomain_result,tab:joint_geo_motion},
we observe that these pairwise approaches usually exhibit degraded performance when extended to video sequences, while our method outperforms state-of-the-art methods by 38.64\% in geometry and 25.0\% in motion on average.
Note that,
unlike Zero-MSF~\cite{liang2025zero},
we do not train our model with motion annotations from Dynamic Replica~\cite{karaev2023dynamicstereo} and Point Odyssey~\cite{zheng2023pointodyssey}, yet still achieve better performance except on one comparable metric.
\Cref{fig:effectiveness} shows that our method estimates temporally consistent scene flow in a world coordinate system, 
remaining robust to camera motion and capable of describing 4D scene dynamics more accurately and efficiently.

\begin{table}[t]
\centering
\caption{
\textbf{Evaluation on world-centric geometric reconstruction.}
% Lower $\text{Rel}^p$ and higher $\delta^p$ indicate better performance.
$^\dagger$denotes using post-optimization. 
Note that,
our results are reported without any post-optimization.
}
\vspace{-0.5em}
\label{tab:geo_only}
\begin{adjustbox}{width=\linewidth}
\begin{tabular}{@{}l cc cc cc c@{}}
\toprule
\multirow{2}{*}{\textbf{Method}} &
\multicolumn{2}{c}{\textbf{Monkaa}~\cite{mayer2016large}} &
\multicolumn{2}{c}{\textbf{Sintel}~\cite{sintel}} &
\multicolumn{2}{c}{\textbf{DDAD}~\cite{ddad2020}} &
\textbf{Avg.}  \\
\cmidrule(lr){2-3} \cmidrule(lr){4-5} \cmidrule(lr){6-7}
&
$\text{Rel}^{p}\!\downarrow$ & $\delta^{p}\!\uparrow$ &
$\text{Rel}^{p}\!\downarrow$ & $\delta^{p}\!\uparrow$ &
$\text{Rel}^{p}\!\downarrow$ & $\delta^{p}\!\uparrow$ &
\textbf{Rank} $\!\downarrow$
\\
\midrule
\ \underline{Camera-centric} & & & & & & & \\ 
DepthPro~\cite{bochkovskiy2025depth} & 36.96 & 63.65 & 43.30 & 42.30 & 35.38 & 45.49 & 7.00 \\
MoGe~\cite{wang2024moge} & 35.21 & 65.95 & 35.28 & 60.97 & 18.63 & 77.07 & 4.33 \\
GeoCrafter~\cite{xu2025geometrycrafter} & 33.44 & 65.79 & 30.61 & 68.07 & 19.17 & 76.39 & 3.33 \\
\midrule
\ \underline{World-centric} & & & & & & & \\
MonST3R~\cite{zhang24monst3r}$^\dagger$ & 41.41 & 31.46 & 37.65 & 51.41 & 31.56 & 55.30 & 6.33 \\
VGGT~\cite{wang2025vggt} & 34.54 & 56.65 & \textbf{26.83} & \textbf{67.91} & 15.98 & 84.06 & \textbf{2.33} \\
Geo4D~\cite{jiang2025geo4d}$^\dagger$ & 28.04 & 69.52 & 34.61 & 59.54 & \textbf{14.58} & \textbf{83.68} & \textbf{2.33} \\
St4RTrack~\cite{st4rtrack2025} & 47.04 & 45.46 & 40.59 & 51.54 & 39.59 & 32.47 & 7.67 \\
\textbf{Ours} & \textbf{25.88} & \textbf{74.01} & 32.46 & 63.14 & 21.27 & 72.82 & 2.67 \\
\bottomrule
\end{tabular}
\end{adjustbox}
\vspace{-1em}
\end{table}

\paragraph{Assessing Geometry Reconstruction}
We further compare our method with several representative approaches in geometry reconstruction in~\cref{tab:geo_only}.
We group them by their geometric representation:
(1) Camera-centric methods predict depth or point maps in the camera coordinate system.
We transform their outputs to world coordinates using VGGT poses for fair comparison.
(2) World-centric methods predict geometry directly in the reference frame coordinate system, and we align their results to ground truth using an affine transformation.
Our method achieves \emph{state-of-the-art} performance on Monkaa, 
demonstrating the advantages of our proposed architecture. 
For Sintel and DDAD, our performance is inferior to VGGT~\cite{wang2025vggt}, 
which we attribute to our single-modal design (w/o camera rays and depth maps) and the limited scale of our outdoor training data. 
Note that we do not perform post-optimization, as in Geo4D~\cite{jiang2025geo4d}.
The visual comparisons (see in supp.) further validate that our approach produces more coherent and consistent reconstructions in dynamic environments.

\subsection{Ablation Study}
We conduct thorough ablations to analyze \method.
Results are reported in~\cref{tab:ablation_geometry,tab:ablation_motion},
aiming to investigate the following three key questions:

\begin{table}[t]
\centering
\caption{
\textbf{Ablation study on geometry VAE.}
Here,
we report geometry reconstruction results on both VAE and U-Net,
as a better VAE may not always lead to better final results.
The models are trained on a subset only for geometry reconstruction.
}
\label{tab:ablation_geometry}
\vspace{-0.5em}
\begin{adjustbox}{width=\linewidth}
\begin{tabular}{@{}l cc cc cc@{}}
\toprule
\multirow{2}{*}{\textbf{Model}} & 
\multirow{2}{*}{\textbf{Training Type}} & 
\multirow{2}{*}{\textbf{Rescale}} & 
\multicolumn{2}{c}{\textbf{Sintel}\cite{sintel}} &
\multicolumn{2}{c}{\textbf{Monkaa}~\cite{mayer2016large}} \\
\cmidrule(lr){4-5} \cmidrule(lr){6-7}
& & &
$\text{Rel}^{p}\!\downarrow$ & $\delta^{p}\!\uparrow$ &
$\text{Rel}^{p}\!\downarrow$ & $\delta^{p}\!\uparrow$ \\
\midrule
VAE - 1 & Original  & Max &
39.96 & 62.63 &
23.78 & 67.33 \\

VAE - 2 & From scratch & Max &
18.80 & 79.68 &
11.48 & 90.55 \\

VAE - 3  & Finetune decoder & Max &
20.76 & 77.77 &
14.44 & 85.91 \\

VAE - 4 & Finetune all & Mean &
\textbf{5.68} & \textbf{98.77} &
\textbf{5.03} & \textbf{99.13} \\

\midrule
Unet - I & VAE - 3 + Unet & Max &  40.47 & 49.42 & 33.66 & 56.42  \\
Unet - II & VAE - 4 + Unet & Mean &  \textbf{35.17} & \textbf{58.08} & \textbf{27.36} & \textbf{66.21} \\

\bottomrule
\end{tabular}
\end{adjustbox}
\vspace{-0.5em}
\end{table}

\noindent
\textbf{Is it necessary to strictly align the input distribution with that of Video Diffusion?}
The answer is \textbf{no}.
Most of the existing geometric diffusion models~\cite{ke2024repurposing,zhang2024world,jiang2025geo4d}
strictly normalize the 3D attributes (such as depth and point maps) to $[-1, 1]$,
in order to inherit the diffusion prior.
However,
we find that such a \textit{max-rescale} normalization results in suboptimal reconstruction accuracy for the pretrained VAEs (VAE-1\&2 of~\cref{tab:ablation_geometry}). 
Although Geo4D~\cite{jiang2025geo4d} alleviates this issue by freezing the VAE encoder and fine-tuning only the decoder, 
our ablation shows that this strategy still yields inferior results (VAE-3 in~\cref{tab:ablation_geometry}).
In contrast,
our proposed \emph{mean rescale} strategy---combined with fine-tuning all VAE components---achieves the best performance without strict adherence to the original VAE distribution (VAE-4 of~\cref{tab:ablation_geometry}). 
We further validate this finding in the Diffusion Unet stage (Unet-I vs. Unet-II),
resulting in an average of 16.6\% gain in geometry.
This finding suggests that \method maintains a better generalization ability,
even do \emph{not} align the distribution with that of video diffusion.

\begin{table}[t]
\centering
\caption{\textbf{Ablation study on motion VAE}.
Comparison of different designs across three dynamic scene flow datasets.
Again, we report results on both VAE and U-Net.
}
\label{tab:ablation_motion}
\vspace{-0.5em}
\begin{adjustbox}{width=\linewidth}
\begin{tabular}{@{}l c cc cc@{}}
\toprule
\multirow{2}{*}{\textbf{Model}} & 
\multirow{2}{*}{\textbf{Fusion Type}} & 
\multicolumn{2}{c}{\textbf{Spring}~\cite{Mehl2023_Spring}} &
\multicolumn{2}{c}{\textbf{Point Odyssey}~\cite{zheng2023pointodyssey}} \\
\cmidrule(lr){3-4} \cmidrule(lr){5-6}
 &  & EPE$\downarrow$ & APD$_{0.03}\uparrow$ & EPE$\downarrow$ & APD$_{0.03}\uparrow$ \\
\midrule
VAE - 5 & Original &  6.43 & 50.49 & 1.55 & 91.68 \\
VAE - 6 & Offset &  1.83 & 83.51 & 2.00 & 86.66 \\
VAE - 7 & Separate &  \textbf{0.66} & \textbf{96.75} & \textbf{0.77} & \textbf{94.74} \\
VAE - 8 & Unify &  0.88 & 94.78 & \textbf{0.77} & 94.19 \\
\midrule
Unet -III & Separate & 6.37 & 65.94 & 3.65 & 63.19 \\
Unet - IV & Unify & \textbf{5.16} & \textbf{72.81} & \textbf{3.49} & \textbf{66.36} \\
\bottomrule
\end{tabular}
\end{adjustbox}
\vspace{-0.5em}
\end{table}

\noindent
\textbf{Which strategy for fusing geometric and motion latent information is most effective?}
We explore different strategies,
including \emph{Offset}, \emph{Separate}, and \emph{Unify} as discussed in~\cref{sec:method_vae},
for jointly encoding geometry and motion information in~\cref{tab:ablation_motion}.
Experimental results indicate that although separate VAEs achieve optimal performance in VAE reconstruction, a unified VAE ultimately performs better in Unet prediction.
This phenomenon highlights the importance of tightly coupling geometry and motion representations for coherent 4D modeling.

\noindent
\textbf{What prior knowledge does video generator provide?}
When using the original pretrained VAE to encode geometry and motion,
we observe that the model already exhibits reasonable reconstruction ability in indoor scenes, as seen in the first rows of~\cref{tab:ablation_geometry,tab:ablation_motion}. 
However, due to the significant scale discrepancy between point maps, scene flow, and image-space distributions,
the original VAE struggles to handle large-scale variations, especially in outdoor scenes, shown in~\cref{fig:rescale}. 
We argue that appropriate normalization and fine-tuning strategies are essential to fully leverage the priors embedded in the diffusion model. 
As shown in VAE-2 of Table~\ref{tab:ablation_geometry}, 
training from scratch leads to suboptimal results, confirming that the pretrained Video Diffusion model indeed contains rich priors beneficial for dense 4D reconstruction. 
By explicitly modeling these priors, we effectively unlock the 4D representation capability of the video diffusion model.

\section{Conclusion}
\label{sec:conclusion}

We introduce \method,
a framework that is capable of jointly reconstructing dense geometry and motion from a monocular video.
By defining both in a unified world coordinate system
and designing a novel 4D VAE that encodes them into a shared latent space,
we achieve state-of-the-art performance even without any post-processing. 
Notably,
we show that it is not necessary to strictly align the 4D latent distribution with the original SVD latent distribution.
In fact,
our relaxed alignment and VAE retraining strategy not only preserves but also improves the diffusion model's generalization,
offering broader insight into adapting diffusion priors to new modalities.

\paragraph{Limitations}
Currently,
we focus solely on dense geometry and motion reconstruction,
but prior work has shown that incorporating multiple geometric modalities can substantially improve the prediction of 3D attributes, including camera parameters, point maps, depth maps, point tracks, and novel views.
Thus, exploring multimodal integration is a promising direction for future work.

\section*{Acknowledgments}
% Ruijie Zhu completed this work during his internship at Tencent ARC Lab. We thank Tian-Xing Xu for providing the codebase of GeometryCrafter.
Xiaoguang Han is supported by Guangdong Provincial Outstanding Youth Fund with No. 2023B1515020055.
Chuanxia Zheng is supported by NTU SUG-NAP and the National Research Foundation, Singapore, under its NRF Fellowship Award NRF-NRFF17-2025-0009.
\\
% \clearpage
{
    \small
    \bibliographystyle{ieeenat_fullname}
    \bibliography{main,chuanxia_specific,chuanxia_general}
}

\clearpage
\setcounter{section}{0}
\maketitlesupplementary

\renewcommand{\thesection}{\Alph{section}}
\renewcommand{\thesubsection}{\thesection.\arabic{subsection}}
\renewcommand{\thesubsubsection}{\thesubsection.\arabic{subsubsection}}

In the \textbf{supplementary video}, we provide additional visual results.
In this \textbf{supplementary document}, we present further details and analyses to complement the main paper.

\section{Data Processing}
\label{sec:preprocess}

For each video sequence,
we preprocess the corresponding point maps and scene flow into a unified [world]-coordinate system,
referenced by the first camera pose.
The processing pipeline consists of three steps:
(1) camera-pose normalization,
(2) transformation of point maps and scene flow into the world coordinate frame, and
(3) global normalization of world-space geometry and motion. Below, we detail each component.

% ---------------------------
\subsection{Camera Pose Normalization}

Monocular reconstruction systems often produce camera poses that contain arbitrary global rotation and translation.
To eliminate this ambiguity,
following DUSt3R~\cite{wang2024dust3r},
we align all poses to a canonical coordinate frame defined by the first camera.
In particular,
given a sequence of camera poses 
$\{\bm{P}_i\}_{i=1}^N$,
where each $\bm{P}_i \in \mathbb{R}^{4 \times 4}$, 
we decompose the first pose as
\begin{equation}
\bm{R}_0 = \bm{P}_0[:3,:3], \qquad \bm{t}_0 = \bm{P}_0[:3,3].
\end{equation}
Each pose is then normalized by
\begin{equation}
\tilde{\bm{R}}_i = \bm{R}_0^\top \bm{R}_i, \qquad
\tilde{\bm{t}}_i = \bm{R}_0^\top (\bm{t}_i - \bm{t}_0),
\end{equation}
which preserves the relative motion within the sequence while removing global rotation and translation.

% ---------------------------
\subsection{Point Map Transformation}

Given a point map $\bm{X}_i^C \in \mathbb{R}^3$ expressed in the camera coordinate system of frame $i$, we transform it into the first-frame coordinate system using the normalized camera poses:
\begin{equation}
\bm{X}_i = \tilde{\bm{R}}_i \bm{X}_i^C + \tilde{\bm{t}}_i .
\end{equation}
This transformation is applied to all valid pixels, while invalid pixels (as indicated by the validity mask) are set to zero. 
For these invalid points, we use pyramid padding~\cite{zhang2024layerdiffuse} to fill them in.
Note that,
we do not supervise these invalid points; filling them in is solely to prevent the VAE from being affected by missing values during feature extraction.

% ---------------------------
\subsection{Scene Flow Transformation}

The original scene flow $ \bm{V}_i^C$ is defined in the camera coordinates of frame $i$. To obtain the first-frame (world) space scene flow, we first compute the deformed points:
\begin{equation}
\bm{X}_{i \rightarrow i+1}^{C} = \bm{X}_i^C + \bm{V}_i^C,
\end{equation}
and transform them using the camera pose of the next frame:
\begin{equation}
\bm{X}_{i \rightarrow i+1} = 
\tilde{\bm{R}}_{i+1} \bm{X}_{i \rightarrow i+1}^{C} + \tilde{\bm{t}}_{i+1}.
\end{equation}
The world-space scene flow is computed as
\begin{equation}
\bm{V}_i = \bm{X}_{i \rightarrow i+1} - \bm{X}_i .
\end{equation}
If a deformability mask is available, we apply it to zero out scene flow in non-dynamic regions.

% ---------------------------
\subsection{Global World-Coordinate Normalization}

To ensure consistent training across scenes with different scales,
the global normalization is applied in the [world]-space geometry.

\paragraph{Centering}
We first compute the centroid of all valid points:
\begin{equation}
\bm{\mu} = \frac{1}{| \mathcal{D}|} \sum_{d \in \mathcal{D}} \bm{X}_d.
\end{equation}

\paragraph{Isotropic Rescaling}
Instead of scaling by the maximum radius, we compute the mean scale of valid points to the centroid:
\begin{equation}
S = \frac{1}{\mathcal{D}} \sum_{d \in \mathcal{D}} \lVert \bm{X}_d -\bm{\mu} \rVert_2.
\end{equation}
Then we uniformly normalize the point map, camera pose, and scene flow with the affine transformation:
\begin{equation}
\bm{X}_i \leftarrow \frac{\bm{X}_i - \bm{\mu}}{S}, \qquad
\tilde{\bm{t}}_i \leftarrow \frac{\tilde{\bm{t}}_i - \bm{\mu}}{S}, \qquad
\bm{V}_i \leftarrow \frac{\bm{V}_i}{S}.
\end{equation}
This isotropic scaling preserves the geometric structure while normalizing the absolute scale across datasets.
The normalization parameters $(\bm{\mu}, S)$ are stored for optional recovery of the original metric scale.

\section{Additional Ablations}
\label{sec:more_ablations}

\begin{table*}[t]
\centering
\caption{
\textbf{Ablation study on Geometry VAE components.}
Metrics are reported for \textbf{ScanNet}, \textbf{Sintel}, and \textbf{Monkaa} datasets:
point accuracy ($\text{Rel}^{p}\!\downarrow$, $\delta^{p}\!\uparrow$) and depth accuracy ($\text{Rel}^{d}\!\downarrow$, $\delta^{d}\!\uparrow$).
}
\label{tab:more_ablation_vae}
\begin{adjustbox}{width=\textwidth}
\begin{tabular}{cccc|cccc|cccc|cccc}
\toprule
\multirow{2}{*}{\textbf{Model}} & 
\multirow{2}{*}{\textbf{Training}} & 
\multirow{2}{*}{\textbf{Rescale}} & 
\multirow{2}{*}{\textbf{Depth Loss}} &
\multicolumn{4}{c|}{\textbf{ScanNet}} &
\multicolumn{4}{c|}{\textbf{Sintel}} &
\multicolumn{4}{c}{\textbf{Monkaa}} \\
% \cmidrule(lr){5-8} \cmidrule(lr){9-12} \cmidrule(lr){13-16}
& & & &
$\text{Rel}^{p}\!\downarrow$ & $\delta^{p}\!\uparrow$ &
$\text{Rel}^{d}\!\downarrow$ & $\delta^{d}\!\uparrow$ &
$\text{Rel}^{p}\!\downarrow$ & $\delta^{p}\!\uparrow$ &
$\text{Rel}^{d}\!\downarrow$ & $\delta^{d}\!\uparrow$ &
$\text{Rel}^{p}\!\downarrow$ & $\delta^{p}\!\uparrow$ &
$\text{Rel}^{d}\!\downarrow$ & $\delta^{d}\!\uparrow$ \\
\midrule
1 & Original  & Max & $\times$ &
14.96 & 86.95 & 1.98 & 99.90 &
39.96 & 62.63 & 12.62 & 92.65 &
23.78 & 67.33 & 4.30 & 98.91 \\

2 & From scratch & Mean & $\times$ &
7.01 & 99.29 & 1.88 & 99.92 &
10.40 & 93.89 & 42.44 & 92.38 &
8.02 & 96.91 & 4.10 & 98.47 \\

3 & From scratch & Max & $\times$ &
11.88 & 91.08 & 1.82 & 99.59 &
18.80 & 79.68 & 10.21 & 92.26 &
11.48 & 90.55 & 5.93 & 93.80 \\

4 & Finetune decoder & Max & $\times$ &
10.45 & 92.30 & 2.73 & 98.83 &
20.76 & 77.77 & 12.67 & 89.08 &
14.44 & 85.91 & 7.10 & 91.66 \\

5 & Finetune decoder & Max & $\checkmark$ &
4.46 & 98.20 & \textbf{0.54} & 99.97 &
12.04 & 88.28 & \textbf{4.30} & \textbf{98.64} &
8.50 & 93.36 & \textbf{1.37} & 99.64 \\

6 & Finetune all & Mean & $\times$ &
4.46 & 99.69 & 1.11 & 99.97 &
5.68 & 98.77 & 9.73 & 94.60 &
5.03 & 99.13 & 3.54 & 99.27 \\

7 & Finetune all & Mean & $\checkmark$ &
\textbf{3.03} & \textbf{99.88} & 0.76 & \textbf{99.99} &
\textbf{4.39} & \textbf{99.14} & 8.04 & 95.31 &
\textbf{3.74} & \textbf{99.47} & 1.83 & \textbf{99.77} \\
\bottomrule
\end{tabular}
\end{adjustbox}
\end{table*}

\begin{table*}[t]
\centering
\renewcommand{\arraystretch}{1.2}
\caption{\textbf{Ablation study on Unet components for Geometry Reconstruction.} 
We compare models trained with different strategies, rescaling methods, and decoder losses.}
\label{tab:more_ablation_unet}
\begin{adjustbox}{width=\textwidth}
\begin{tabular}{c|ccc|cc|cc|cc|cc}
\toprule
\multirow{2}{*}{\textbf{Model}} & \multirow{2}{*}{\textbf{Training Strategy}} & \multirow{2}{*}{\textbf{Normalization}} & \multirow{2}{*}{\textbf{Decoder Loss}} & 
\multicolumn{2}{c|}{\textbf{GMU Kitchen}} & 
\multicolumn{2}{c|}{\textbf{Monkaa}} & 
\multicolumn{2}{c|}{\textbf{Sintel}} & 
\multicolumn{2}{c}{\textbf{DDAD}} \\
% \cmidrule(lr){5-12}
 & & & & 
$\text{Rel}^{p}\!\downarrow$ & $\delta^{p}\!\uparrow$ &
$\text{Rel}^{p}\!\downarrow$ & $\delta^{p}\!\uparrow$ &
$\text{Rel}^{p}\!\downarrow$ & $\delta^{p}\!\uparrow$ &
$\text{Rel}^{p}\!\downarrow$ & $\delta^{p}\!\uparrow$  \\
\midrule
I & Finetuning VAE Decoder & Max & \(\times\) & 17.49 & 80.65 & 33.66 & 56.42 & 40.47 & 49.42 & 41.66 & 39.45 \\
II & Finetuning the whole VAE & Mean & \(\times\) & 17.72 & 82.90 & 27.36 & 66.21 & 35.17 & \textbf{58.08} & 33.78 & 50.34 \\
III & Finetuning the whole VAE & Mean & \(\checkmark\) & \textbf{14.47} & \textbf{90.54} & \textbf{25.33} & \textbf{71.77} & \textbf{33.39} & 56.94  & \textbf{22.39} & \textbf{70.42} \\
\bottomrule
\end{tabular}
\end{adjustbox}
\end{table*}

\subsection{Ablation on the Multimodal Supervision}

\paragraph{Motivation}
While methods such as Geo4D~\cite{jiang2025geo4d} rely on multi-modality outputs 
(e.g., depth, point maps, and normals) together with a post-optimization fusion stage 
to obtain the final reconstruction,
our main goal is to achieve 
\emph{fully feed-forward} 4D geometry and motion reconstruction.
Therefore, we intentionally avoid introducing any auxiliary outputs or post-refinement during inference.
Interestingly, although multimodal outputs are not used at test time,
we find that multimodal \emph{supervision} during VAE training can still benefit the reconstruction quality of the 4D latent.
In particular, we leverage depth as an additional supervision signal derived from the world-coordinate point maps.

\paragraph{Depth supervision}
Given the reconstructed world-coordinate point map $\hat{\mathbf{X}}$ and the ground-truth point map $\mathbf{X}$,
we project both into the depth domain using the normalized camera pose $\tilde{\mathbf{P}}$:
\begin{equation}
    \hat{\bm{D}} = \Pi(\hat{\bm{X}}, \tilde{\bm{P}}), \qquad 
    \bm{D} = \Pi(\bm{X}, \tilde{\bm{P}}),
\end{equation}
where $\Pi(\cdot)$ denotes standard projection into the depth map.
We apply two complementary losses:

\begin{itemize}
    \item \textbf{Per-pixel L1 depth loss}.  
    This loss encourages accurate depth prediction and is masked by the depth validity map:
    \begin{equation}
        \mathcal{L}_{\text{L1-D}} = 
        \left\|
        \, (\hat{\mathbf{D}} - \mathbf{D}) \odot \mathbf{W}
        \right\|_{1},
    \end{equation}
    where $\mathbf{W}$ is the binary valid-mask.

    \item \textbf{Multi-scale patch depth loss}.  
    To improve geometric consistency across different spatial scales, we compute an L1 loss over patches defined by the scale factors $\{4,16,64\}$.
    For each scale, the depth maps are divided into non-overlapping patches; within each patch,  
    the mean depth (computed with masked averaging) is subtracted to remove global bias:
    \begin{equation}
        \mathcal{L}_{\text{Patch-D}} 
        = 
        \sum_{s \in \{4,16,64\}}
        \left\|
        \, \left( \hat{\mathbf{D}}^{(s)} - \mathbf{D}^{(s)} \right)
        \odot \mathbf{W}^{(s)}
        \right\|_{1}.
    \end{equation}
    This term encourages consistent local geometric structure and suppresses depth-shift artifacts.
\end{itemize}

The total multimodal depth supervision is:
\begin{equation}
    \mathcal{L}_{\text{depth}} 
    = 
    \lambda_{\text{L1-D}} \mathcal{L}_{\text{L1-D}}
    +
    \lambda_{\text{Patch-D}} \mathcal{L}_{\text{Patch-D}}.
\end{equation}  

\paragraph{Results}
As shown in~\cref{tab:more_ablation_vae}, introducing depth-based multimodal supervision 
significantly improves the reconstruction quality of the world-coordinate point maps (with 13.55\% improvement in point map and 16.41\% in depth map).
Notably, this improvement is achieved \emph{without} modifying the inference pipeline or introducing any extra modalities at test time.
This ablation demonstrates that multimodal supervision is an effective strategy for enhancing the 4D latent representation learned by our VAE.

\subsection{Ablation on the Decoder Loss}

\paragraph{Motivation}
In the deterministic setting, the denoising process in conventional diffusion models can be viewed as collapsing from a multi-step procedure into a single step. 
Therefore, in addition to supervising the denoised latent representation, we introduce a \emph{decoder loss} that directly supervises the VAE decoder's output. 
Compared to latent regression, this supervision is more direct and provides stronger training signals to the UNet.

\paragraph{Implementation}
During training, we do not update the VAE decoder's weights. 
However, we still compute its gradients so that the loss can be back-propagated through the decoder to update the UNet parameters. 
To reduce memory consumption, we apply gradient checkpointing to the VAE decoder during this process.

\paragraph{Results}
As shown in~\cref{tab:more_ablation_unet}, incorporating the decoder loss consistently improves the UNet training. 
Across four unseen datasets, it yields an average improvement of {15.01\%}, with particularly notable gains on the outdoor dataset DDAD, where the performance improves by {36.80\%}.

\subsection{Ablation on the training paradigm}

\paragraph{Motivation}
Following prior works~\cite{hu2024-DepthCrafter,xu2025geometrycrafter} in employing EDM~\cite{karras2022elucidating} pre-conditioning, our framework supports both the \textit{deterministic} and \textit{denoising} diffusion paradigms, on top of the pretrained SVD model.
Since 4D Reconstruction is a deterministic task, we use a deterministic paradigm by default, which has been widely explored and shown to be effective in previous dense prediction frameworks~\cite{xu2024matters, song2025depthmaster, xu2025geometrycrafter}.
However, we also want to know exactly how different these two training paradigms are in our framework, especially for 4D latents that incorporate both geometry and motion information. Therefore, we conduct ablation experiments to verify this.

\paragraph{Implementation}
As described in~\Cref{sec:method_training}, we have already introduced the loss functions for two training paradigms. For a fair comparison, we do not use decoder loss in the deterministic paradigm. Specifically, we directly feed video latents into the U-Net to predict 4D latents. For the denoising paradigm, we first add noise to the 4D latents and channel-wise concatenate the video latents, then use the U-Net's multi-step denoising to predict the 4D latents. All U-Net weights are initialized from the original SVD, with channel dimensions adjusted only at the first layer to accommodate different training paradigms.

\paragraph{Results}
As shown in~\cref{tab:diffusion_vs_determ},  the deterministic paradigm reduces $\text{Rel}^{p}$ by about 12.4\% and improves $\delta^{p}$ by approximately 12.7\% compared to the diffusion paradigms averaged across datasets. 
This result strongly demonstrates the effectiveness of the deterministic paradigm in dense prediction tasks. Also, this shows that prior knowledge of SVD can be inherited by the model without relying on a denoising mechanism.

\begin{table}[h]
% \vspace{-1em}
\centering
\caption{
\textbf{Ablation on different training paradigm.}
}
\label{tab:diffusion_vs_determ}
% \vspace{-0.5em}
\begin{adjustbox}{width=\linewidth}
\begin{tabular}{@{}l cc cc cc@{}}
\toprule
\multirow{2}{*}{\textbf{Training Type}} & 
\multicolumn{2}{c}{\textbf{Monkaa}} &
\multicolumn{2}{c}{\textbf{Sintel}} &
\multicolumn{2}{c}{\textbf{DDAD}} \\
\cmidrule(lr){2-3} \cmidrule(lr){4-5} \cmidrule(lr){6-7}
& 
$\text{Rel}^{p}\!\downarrow$ & $\delta^{p}\!\uparrow$ &
$\text{Rel}^{p}\!\downarrow$ & $\delta^{p}\!\uparrow$ &
$\text{Rel}^{p}\!\downarrow$ & $\delta^{p}\!\uparrow$ \\
\midrule
Diffusion &  {30.11} & {65.49} & {35.95} & {53.82} & 24.58 & 67.54 \\
Deterministic &  \textbf{25.88} & \textbf{74.01} & \textbf{32.46} & \textbf{63.14} & \textbf{21.27} & \textbf{72.82} \\
\bottomrule
\end{tabular}
\end{adjustbox}
% \vspace{-1em}
\end{table}

\section{Implementation Details}
\label{sec:more_details}

\subsection{Hyperparameter}

We list the loss weights used for training the 4D VAE. 
\begin{itemize}
    \item {Point Map Reconstruction Loss:} 
    $\lambda_{\text{point}} = 1.0$
    \item {Per-pixel L1 Depth loss:} 
    $\lambda_{\text{L1-D}} = 1.0$
    \item {Multi-Scale Depth Supervision:} 
    $\lambda_{\text{Patch-D}} = 1.0$
    \item {Normal Consistency Loss:}
    $\lambda_{\text{normal}} = 0.2$    
    \item {Scene Flow Reconstruction Loss:} 
    $\lambda_{\text{sceneflow}} = 1.0$
    \item {Scene Flow Regulation Loss:} 
    $\lambda_{\text{reg}} = 0.01$
\end{itemize}
The pretrained video diffusion UNet is optimized with the following latent regression loss and decoder loss (optional):

\begin{itemize}
    \item {Latent Regression Loss:} 
    $\lambda_{\text{latent}} = 1.0$
    \item {Point Map Decoder Loss:} 
    $\lambda_{G} = 1.0$
    \item {Scene Flow Decoder Loss:} 
    $\lambda_{M} = 1.0$    
\end{itemize}

\subsection{Used Training Set}
\begin{figure*}
    \centering
    \includegraphics[width=\linewidth]{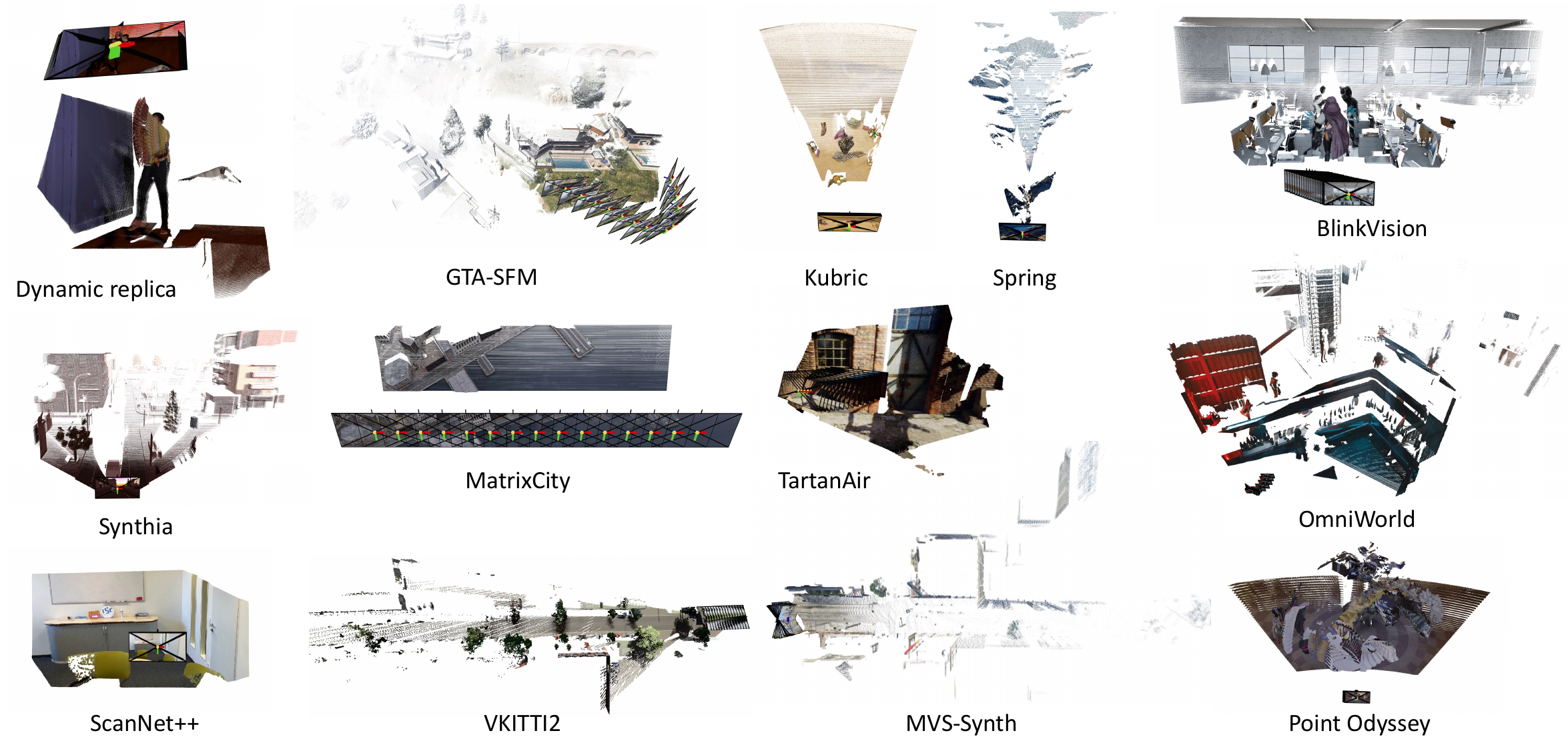}
    \caption{\textbf{The examples of our training set.} We randomly sample video frames from these datasets. In geometric training, we set a random stride to sample the video at different intervals. In motion training, we always keep the stride at 1 to continuously sample frames.}
    \label{fig:trainingset}
\end{figure*}

We list the used training datasets in~\cref{tab:train_dataset}, and provide some visual samples in~\cref{fig:trainingset}.
\begin{table}
\setlength{\tabcolsep}{2pt}
\begin{center}
\caption{\textbf{An overview of the training datasets.} To balance the training, we sample the subset of some datasets.}
\label{tab:train_dataset}
\resizebox{1.0\linewidth}{!}{
\begin{tabular}{cccc}
\hline 
Dataset & Domain & \#Frames & \#Videos\\
\hline
DynamicReplica~\cite{karaev2023dynamicstereo} & Indoor/Outdoor & 145K & 1126 \\
GTA-SfM~\cite{wang2020gtasfm} & Outdoor & 19K & 234 \\
Kubric~\cite{greff2021kubric} & Indoor & 137K & 5736\\
MatrixCity~\cite{li2023matrixcity} & Outdoor-Driving & 452K & 3029 \\ 
MVS-Synth~\cite{huang2018deepmvs} & Outdoor-Driving & 12K & 120 \\
Spring~\cite{Mehl2023_Spring} & Outdoor & 5K & 49 \\
Point Odyssey~\cite{zheng2023pointodyssey} & Indoor & 18K & 120 \\
Synthia~\cite{ros2016synthia} &  Outdoor-Driving & 178K & 1276 \\
TartanAir~\cite{wang2020tartanair} & Outdoor & 306K & 2245 \\
VirtualKitti2~\cite{cabon2020virtual} & Driving & 43K & 320 \\ 
BlinkVision~\cite{li2024blinkvision} & Indoor/Outdoor & 11K & 72 \\
OmniWorld~\cite{zhou2025omniworld} & Indoor/Outdoor & 35K & 350 \\
Scannet++~\cite{yeshwanth2023scannet++} & Indoor-Real & 310K & 2078 \\
\hline
Total & - & 1.67M & 16.8K \\
\hline
\end{tabular} }
\end{center} 
% \vspace{-0.3cm}
\end{table}

\subsection{Model Information}

Our system adopts the VAE and video UNet backbone from the 
stable video diffusion (SVD)~\cite{blattmann2023stable} model.
The pipeline consists of three major components: 
(1) a video VAE encoder for encoding per-frame latent representations,
(2) a 4D VAE decoder for reconstructing geometry and motion fields from latent space,
and (3) a 3D spatiotemporal UNet for latent denoising.
We report the parameter counts of each component below:
\begin{itemize}
    \item \textbf{Video UNet:} 1524.62M parameters.  
    This large spatiotemporal UNet is responsible for denoising the latent
    representations over both space and time, enabling the modeling of dynamic
    geometry and motion.

    \item \textbf{Video VAE Encoder:} 34.16M parameters.  
    This module processes each input frame independently and encodes it into
    a latent space with a spatial downsampling factor of $8\times$.

    \item \textbf{4D VAE Decoder:} 99.00M parameters.  
    This decoder reconstructs 4D point maps and scene flow from the latent
    representation.

    \item \textbf{Total Parameters:} 1657.79M parameters.
\end{itemize}

\paragraph{Inference Timing}
All timings are measured on a single GPU with 40 GB of memory.
For a video clip of {25 frames} at resolution \textbf{$320\times640$}, the average processing time per frame is as follows: 
52.0 ms for VAE encoding, 13.4 ms for latent denoising, and 73.5 ms for VAE
decoding, resulting in a total of 138.9 ms per frame.
These measurements reflect the end-to-end processing required for a full forward
pass of our geometry–motion reconstruction pipeline.

\subsection{Evaluation Metrics}

We provide detailed definitions of the evaluation metrics used for geometry and motion reconstruction.

\paragraph{Geometry Alignment}
Since monocular reconstruction is defined up to scale ambiguity, 
the predicted world-space point map $\hat{\mathbf{X}}_i$ is aligned to the ground truth $\mathbf{X}_i$ 
using a per-sequence scale $s$ and shift $\mathbf{t}$:
\begin{equation}
\tilde{\mathbf{X}}_i = s \hat{\mathbf{X}}_i + \mathbf{t},
\end{equation}
where $s$ and $\mathbf{t}$ are optimized by minimizing:
\begin{equation}
\min_{s,\mathbf{t}} \sum_i 
\left\| s\hat{\mathbf{X}}_i + \mathbf{t} - \mathbf{X}_i \right\|_2^2.
\end{equation}

\paragraph{Relative Point Error ($\mathrm{Rel}^p$)}
We measure the relative geometry error as:
\begin{equation}
\mathrm{Rel}^p =
\frac{1}{N}
\sum_i
\frac{
\left\|\tilde{\mathbf{X}}_i - \mathbf{X}_i\right\|_2
}{
\left\|\mathbf{X}_i\right\|_2
}.
\end{equation}

\paragraph{Inlier Ratio ($\delta^p$)}
We compute the percentage of points whose relative error is below a threshold $\tau$ (0.25 in our experiments):
\begin{equation}
\delta^p =
\frac{1}{N}
\sum_i
\mathbf{1}
\left(
\frac{
\left\|\tilde{\mathbf{X}}_i - \mathbf{X}_i\right\|_2
}{
\left\|\mathbf{X}_i\right\|_2
}
< \tau
\right).
\end{equation}

\paragraph{Scene Flow Alignment}
The predicted scene flow $\hat{\mathbf{V}}_i$ is scaled using the same geometry scale $s$:
\begin{equation}
\tilde{\mathbf{V}}_i = s \hat{\mathbf{V}}_i.
\end{equation}

\paragraph{End-Point Error (EPE)}
We compute the average endpoint error between predicted and ground-truth scene flow:
\begin{equation}
\mathrm{EPE} =
\frac{1}{N}
\sum_i
\left\|
\tilde{\mathbf{V}}_i - \mathbf{V}_i
\right\|_2.
\end{equation}

\paragraph{Average Percent of Points within Delta (APD)}
APD measures the percentage of scene flow vectors whose error is below a threshold $\gamma$:
\begin{equation}
\mathrm{APD}_\gamma =
\frac{1}{N}
\sum_i
\mathbf{1}
\left(
\left\|
\tilde{\mathbf{V}}_i - \mathbf{V}_i
\right\|_2 < \gamma
\right).
\end{equation}

\section{More Visualization Results}
\label{sec:more_vis}

\begin{figure*}
    \centering
    \includegraphics[width=\linewidth]{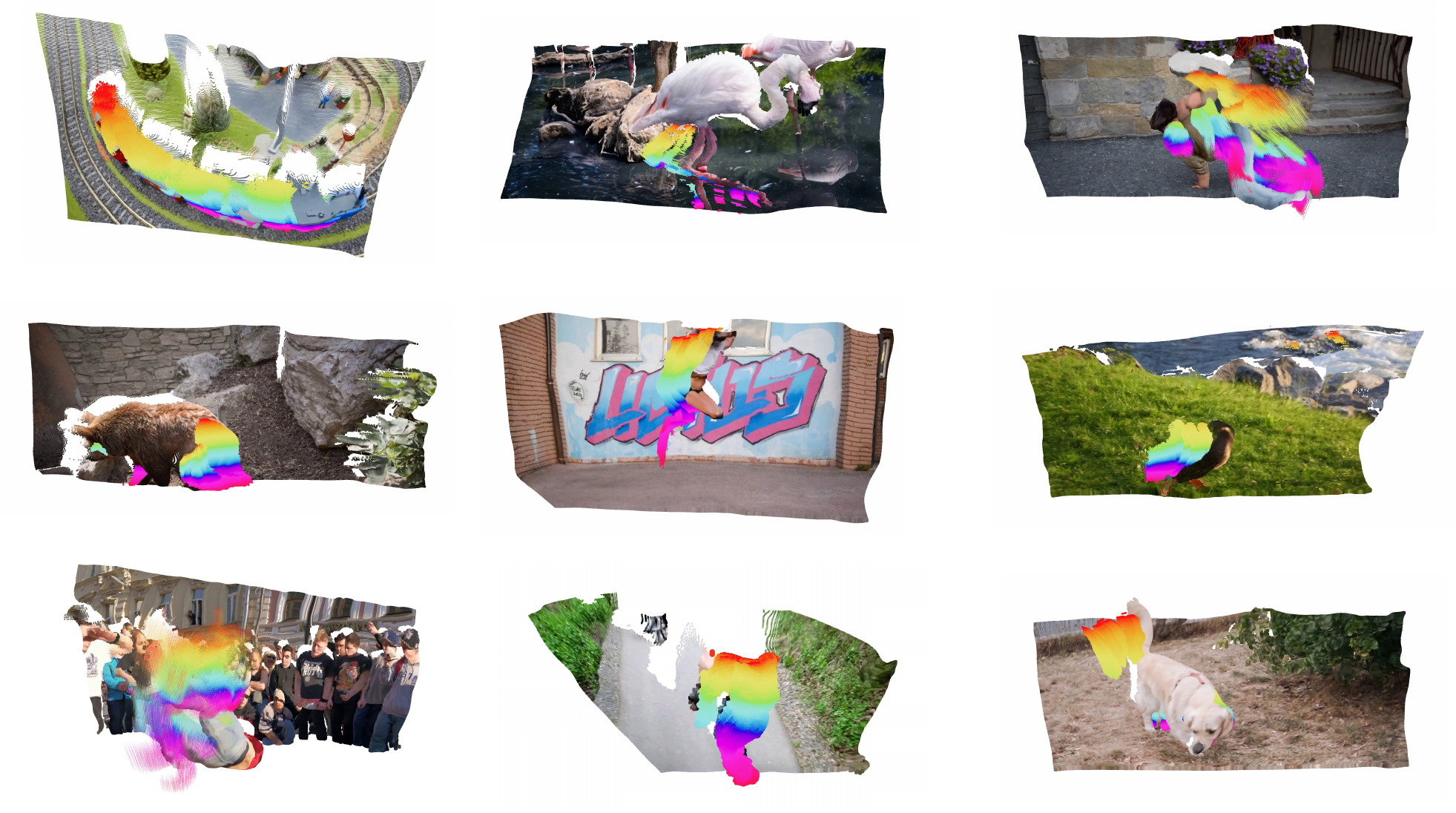}
    \caption{\textbf{Zero-shot results on Davis~\cite{Perazzi2016davis} dataset.} Despite the very limited number of samples used for training scene flow estimation, our method generalizes well across different scene types. Thanks to our end-to-end model design and unified definitions of geometry and motion in the world coordinate system, all results are directly output by the model without any post-optimization. See the video visualization for a more intuitive understanding of the dynamics.}
    \label{fig:zeroshot_davis}
\end{figure*}

\begin{figure*}
    \centering
    \includegraphics[width=0.85\linewidth]{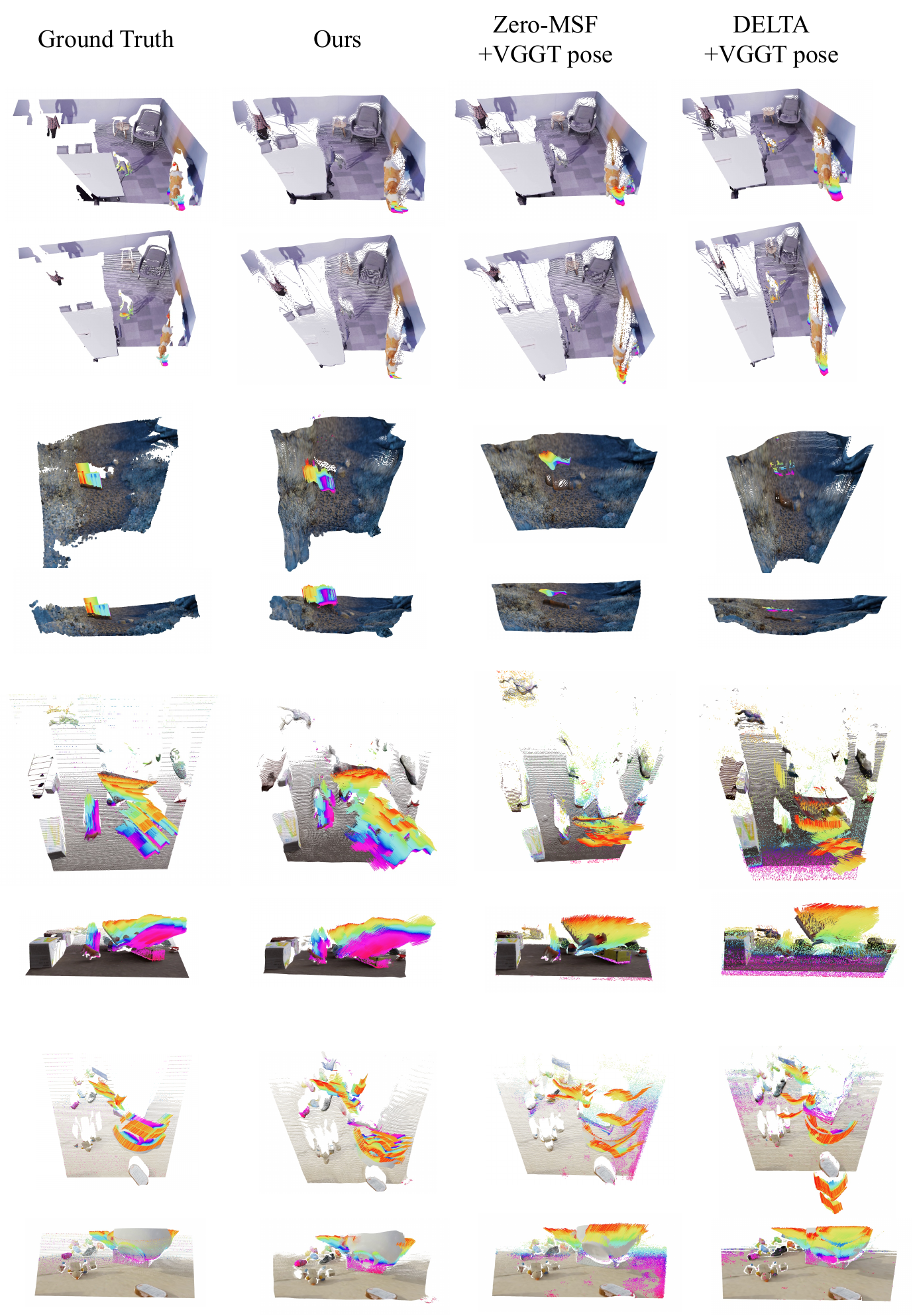}
    \caption{\textbf{Qualitative comparison with the state-of-the-art methods Zero-MSF~\cite{liang2025zero} and DELTA~\cite{ngo2024delta}.} In the first case, our method demonstrates scene flow estimation accuracy comparable to Zero-MSF, even without training on the dynamic replica dataset like it. In the other cases, our method significantly outperforms existing methods in both geometric structure and motion pattern estimation.}
    \label{fig:motion_compare}
\end{figure*}

\begin{figure*}
    \centering
    \includegraphics[width=\linewidth]{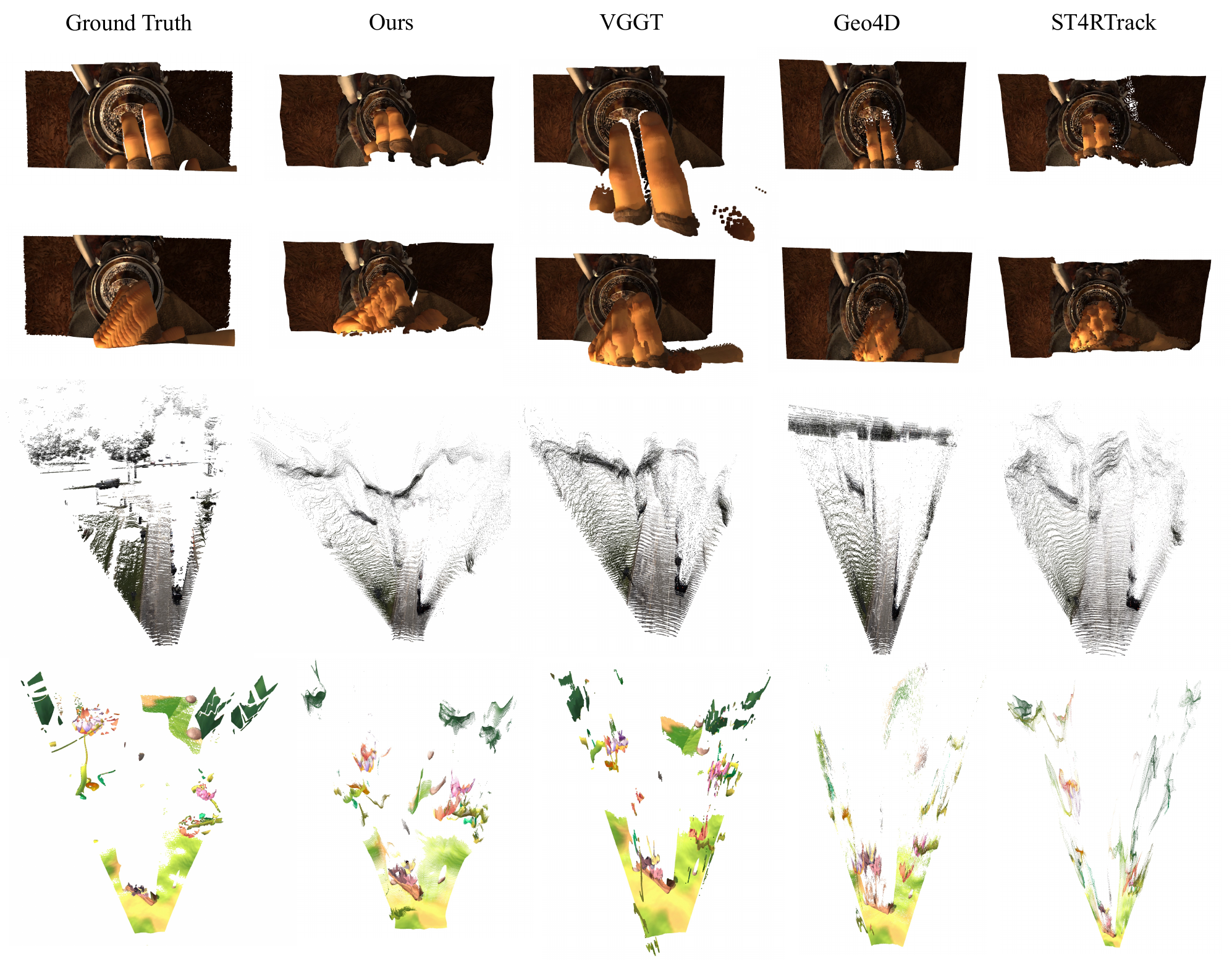}
    \caption{\textbf{Qualitative geometric comparison with VGGT~\cite{wang2025vggt}, Geo4D~\cite{jiang2025geo4d}, and ST4RTrack~\cite{st4rtrack2025}.} For moving objects, such as the finger in the first case, our method estimates more accurate scale and motion changes. For outdoor scenes, our method estimates a more accurate scene structure. Notably, our method, like VGGT, can directly output point clouds in world coordinates without requiring post-optimization steps such as Geo4D. Furthermore, our method has a much smaller training scale than VGGT, yet exhibits good robustness in dynamic scenes. We attribute this to pre-training knowledge of video diffusion and our proposed training strategy.}
    \label{fig:geo_compare}
\end{figure*}

\begin{figure*}[ht]
    \centering
    % \vspace{-1em}
    \includegraphics[width=0.8\linewidth]{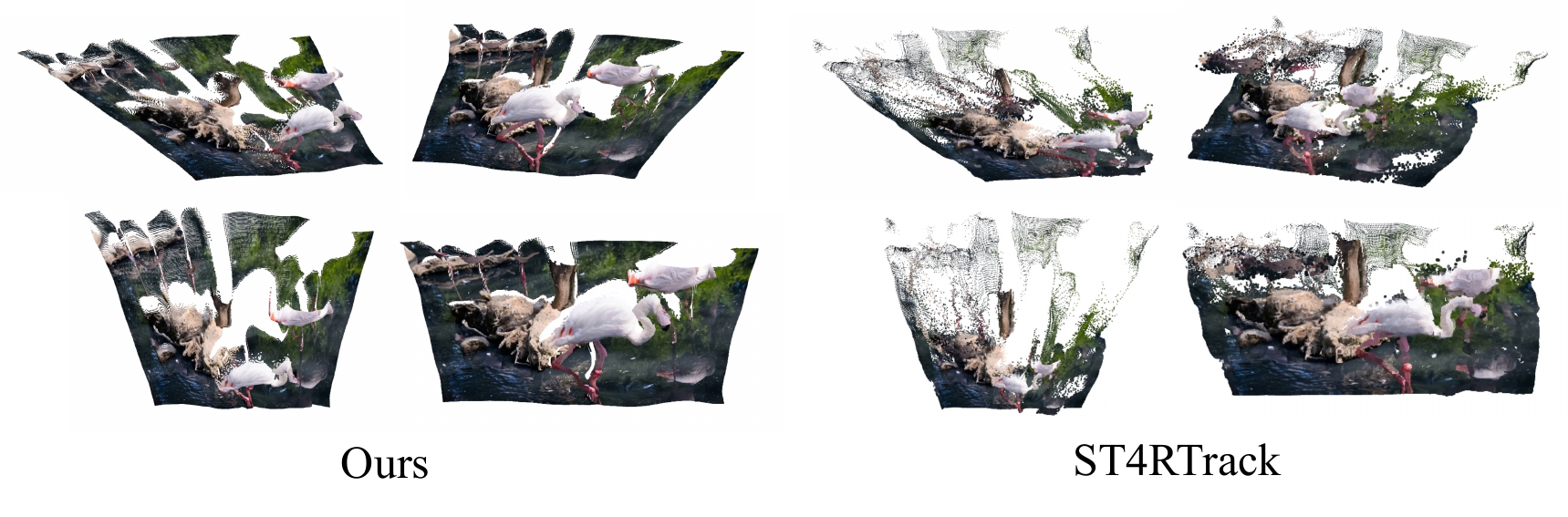}
    % \vspace{-2em}
    \caption{\textbf{Qualitative geometric comparison with ST4RTrack~\cite{st4rtrack2025} on zero-shot generalization.} Compared with ST4RTrack, our results show better multi-view consistency, smoother Geometry, and fewer stray spots.}
    \label{fig:vis_multiview}
    % \vspace{-1em}
\end{figure*}

We select some in-the-wild videos from the Davis~\cite{Perazzi2016davis} dataset as samples for zero-shot testing, and the results are shown in~\cref{fig:zeroshot_davis}.
A more intuitive visualization is provided in the attached video demo.
We also provide more qualitative comparisons with other methods, as shown in~\cref{fig:motion_compare,fig:geo_compare,fig:vis_multiview}.
The comparisons are for two different tasks: 1) joint geometry and motion estimation, and 2) geometry reconstruction only.

\clearpage


\end{document}